\documentclass[sn-mathphys-num]{sn-jnl}

\usepackage{graphicx}%
\usepackage{multirow}%
\usepackage{amsmath,amssymb,amsfonts}%
\usepackage[title]{appendix}%
\usepackage{xcolor}%
\usepackage{textcomp}%
\usepackage{manyfoot}%
\usepackage{booktabs}%
\usepackage{algorithm}%
\usepackage{algorithmicx}%
\usepackage{algpseudocode}%
\usepackage{listings}%
\usepackage{lmodern}
\usepackage{cleveref}
\usepackage{graphicx}
\usepackage{tabularx}
\usepackage{arydshln}
\usepackage{comment}
\usepackage{caption}
\usepackage{subcaption}
\usepackage{soulutf8} 
\usepackage{ifthen}
\usepackage[normalem]{ulem}  
\usepackage{etoolbox} 

\newboolean{showhighlights}
\setboolean{showhighlights}{false} 

\DeclareRobustCommand{\highlight}[1]{%
  \ifthenelse{\boolean{showhighlights}}{%
    \hl{#1}%
  }{%
    #1%
  }%
}

\DeclareRobustCommand{\remove}[1]{%
  \ifthenelse{\boolean{showhighlights}}{%
    \sout{#1}
  }{}%
}

\robustify{\cref}
\robustify{\Cref}
\robustify{\crefrange}
\robustify{\Crefrange}
\robustify{\cite}
\robustify{\citep}
\robustify{\citet}
\soulregister\cite7
\soulregister{\cite}7

\begin{document}

\title[Content-based 3D Image Retrieval and a ColBERT-inspired Re-ranking for Tumor Flagging and Staging]{Content-based 3D Image Retrieval and a ColBERT-inspired Re-ranking for Tumor Flagging and Staging}


\author*[1]{\fnm{Farnaz} \sur{Khun Jush}}\email{farnaz.khunjush@bayer.com}

\author[1]{\fnm{Steffen} \sur{Vogler}}\email{steffen.vogler@bayer.com}

\author[1]{\fnm{Matthias} \sur{Lenga}}\email{matthias.lenga@bayer.com}

\affil[1]{\vspace{-0.5cm}\orgdiv{Radiology R\&D}, \orgname{Bayer AG}, \orgaddress{\street{Müllerstraße 178}, \postcode{13353} \state{Berlin}, \country{Germany}}}


\abstract{\remove{The growing number of medical images in large databases presents significant challenges for radiologists, increasing their workload and complicating the retrieval of diagnostically relevant cases.
Content-based image retrieval (CBIR) systems have the potential to improve clinical outcomes by enabling efficient access to cases that match specific similarity criteria. However, the field is still in its early stages and lacks standardized evaluation methods and comprehensive studies.}
\highlight{The increasing volume of medical images poses challenges for radiologists in retrieving relevant cases. Content-based image retrieval (CBIR) systems offer potential for efficient access to similar cases, yet lack standardized evaluation and comprehensive studies.}
Building on prior \remove{benchmarks}\highlight{studies} for tumor characterization via CBIR, this study advances CBIR research \highlight{for volumetric medical images} through three key contributions: (1) a framework eliminating reliance on pre-segmented data and organ-specific datasets, aligning with large and unstructured image archiving systems, i.e. PACS in clinical practice; (2) \highlight{introduction of C-MIR,} a novel volumetric re-ranking method adapting ColBERT's \highlight{contextualized}  late interaction mechanism \remove{(Contextualized Late Interaction over BERT)}for 3D medical imaging; (3) comprehensive evaluation across \remove{large-scale datasets to assess scalability.}\highlight{four tumor sites using three feature extractors and three database configurations.}
\remove{Our evaluations reveal the strengths and limitations of our proposed re-ranking method, particularly in the context of larger databases. 
Notably, our method effectively localizes the region of interest, rendering the pre-segmentation of datasets obsolete.
}\highlight{Our evaluations highlight the significant advantages of C-MIR. We demonstrate the successful adaptation of the late interaction principle to volumetric medical images, enabling effective context-aware re-ranking. A key finding is C-MIR's ability to effectively localize the region of interest, eliminating the need for pre-segmentation of datasets and offering a computationally efficient alternative to systems relying on expensive data enrichment steps. C-MIR demonstrates promising improvements in tumor flagging, achieving improved performance, particularly for colon and lung tumors ($p<0.05$). C-MIR also shows potential for improving tumor staging, warranting further exploration of its capabilities.}
Ultimately, our work seeks to bridge the gap between advanced retrieval techniques and their practical applications in healthcare, paving the way for improved diagnostic processes.
}



\maketitle

\section{Introduction}\label{sec:introduction}

In the field of computer vision, content-based image retrieval (CBIR) has been extensively studied for decades \cite{dubey2021decade}. Typically, CBIR systems utilize low-dimensional representations \highlight{(embeddings)} of images stored in a database to find similar images based on \remove{distance metrics or}\highlight{embedding} similarity\remove{measures}. Early CBIR methods relied on manually crafted features, which often resulted in loss of important image details due to the constraints of low-dimensional feature design  \cite{dubey2021decade, wang2022two, moirangthem2020content}. However, recent research in deep learning has focused on generating discriminative feature spaces, resulting in more accurate and efficient CBIR methods \cite{qayyum2017medical, dubey2021decade, hu2022x}.
Applying retrieval frameworks to medical images, particularly radiology images, presents ongoing challenges due to the complexity of the task and the nature of medical images, as detailed in \cite{hameed2021content, akgul2011content}. Despite these challenges, \remove{CBIR} the content-based retrieval of medical images offers several advantages\remove{for medical images}, \highlight{e.g.,} enabling radiologists to search for \remove{similar}\highlight{reference} cases and review historical data, reports, patient diagnoses, and prognoses to enhance their decision-making process \cite{hu2022x, hameed2021content, 10966872}. However, \highlight{in} real-world scenarios,\remove{where}\highlight{medical image data is scarcely annotated} and meta-information \highlight{(}such as DICOM \remove{Data}\highlight{headers)} is inconsistent or removed, \highlight{e.g., due to data privacy requirements \cite{portability2012guidance}.} This makes manual searching for relevant images extremely time-consuming and impractical for daily \highlight{clinical} routine work \cite{long2003fundamentals}. Additionally, \highlight{progressing research and development} \remove{of new tools and research }in \highlight{the field of} medical \highlight{imaging} requires \remove{trustworthy}carefully curated, large datasets. Reliable image retrieval method\highlight{s can help to further automate data curation}, making CBIR \highlight{an} essential \highlight{tool} for \remove{advancing computer-aided medical image analysis and diagnosis}\highlight{supporting future advancements in computer-aided medical image analysis and diagnosis} \cite{silva2022computer}.

\highlight{Moreover, while using advanced feature extraction methods has improved the quality of initial retrievals, refining these results to better match clinical relevance remains critical. Re-ranking techniques—which adjust the order of retrieved items using contextual information, user feedback, or advanced similarity metrics—have emerged as a key strategy to enhance precision in CBIR systems \cite{ahmed2020query, khattab2020colbert}. These methods are particularly valuable in medical imaging, where subtle morphological or pathological differences can impact diagnostic decisions \cite{ayadi2018mf,ahmed2020query}.}

Previous research has explored the use of hand-crafted feature extraction techniques for CBIR in medical imaging, with a comprehensive review available in \cite{vishraj2022comprehensive}. More recent studies have highlighted the potential of pre-trained vision embeddings derived from deep neural networks for CBIR in various applications, including anatomical region retrieval for both 2D \cite{sotomayor2021content, denner2024leveraging, lo2024interactive} and 3D images \cite{abacha20233d,10966872,10635170}, near-duplicate detection in radiology \cite{10635550}, as well as pathological tasks \cite{silva2022computer, denner2024leveraging, abacha20233d, mahbod2025evaluating}. 
Notably, the study by \cite{abacha20233d} introduced the first benchmark utilizing these pre-trained embeddings specifically for tumor flagging and staging. Building on \cite{abacha20233d}, we aim to further investigate and refine the application of CBIR in tumor retrieval, addressing the challenges identified in previous studies \highlight{and exploring re-ranking strategies to improve the retrieval results}.

\subsection{Motivation}

Integrating CBIR in tumor retrieval is beneficial for enhancing diagnostic accuracy and efficiency in clinical practices. 
As medical imaging generates vast amounts of data, the ability to swiftly retrieve relevant images based on visual content becomes essential. 
CBIR systems facilitate this by allowing healthcare professionals to quickly access similar cases, thereby improving the decision-making process.
Currently, radiologists often rely on keywords or International Classification of Diseases codes (ICD codes) to locate similar cases within PACS or Radiology Information System (RIS) systems. 
However, this method has limitations. For instance, the search can be refined significantly if images are included as a condition in the search. Moreover, keyword searches can only retrieve scans that were correctly read and labeled initially, meaning that missed pathologies may not surface in these searches. As such, content-based image similarity search becomes a crucial tool for uncovering missed pathologies from historical records, providing a more comprehensive diagnostic approach.
The ability to identify and analyze these missed cases is not only beneficial for patient outcomes but also serves as a valuable feature for quality control departments. The potential to follow up on previously overlooked cases can enhance overall patient care and ensure that health insurance providers are informed of all relevant medical histories.
Moreover, the implementation of CBIR can facilitate research and education by providing access to a diverse range of cases, enriching the training of medical professionals, and fostering a deeper understanding of tumor characteristics and variations.

\subsection{Prior Work}

\subsubsection{CBIR for Tumor Retrieval}

\begin{figure}[!htbp]
  \centering
  \includegraphics[trim=0cm 1.0cm 0.3cm 1cm, clip,width=\textwidth]{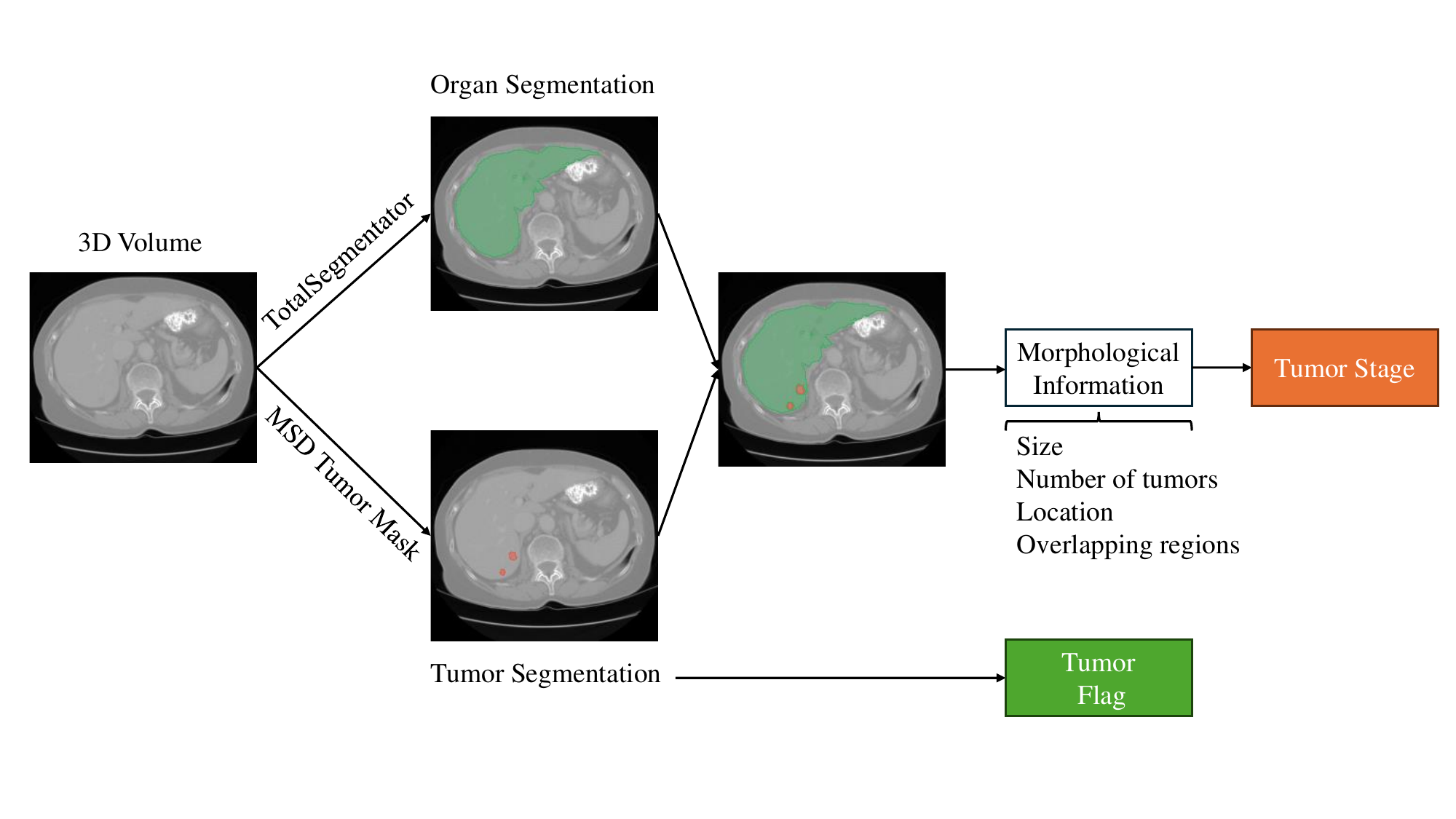}
  \caption{Overview of the \highlight{3D image} processing \remove{of 3D volumes} to extract tumor information based on \cite{abacha20233d}: \remove{The o}\highlight{O}rgan segmentation is performed using the TotalSegmentator model \cite{wasserthal2023totalsegmentator}. The tumor segmentation is taken from the MSD \highlight{ground truth} tumor masks provided by \cite{antonelli2022medical}. The \remove{combined} information from organ segmentation and tumor segmentation \highlight{are used}\highlight{ in combination} to extract morphological details, such as size, number of lesions, location, and overlapping regions. This information is then used to \remove{create}\highlight{derive} tumor stag\remove{ing}\highlight{es} based on the TNM\remove{staging} standard \cite{sobin2009tnm}\remove{, which is later used for database creation and evaluation of retrieval approaches}.}
  \label{fig:overview-dataset}
\end{figure}

In\remove{\cite{zhang2023biomedclip},} \cite{abacha20233d}\remove{proposed} a \remove{benchmark}\highlight{CBIR system} for tumor flagging and staging \remove{for 3D medical image retrieval}\highlight{is proposed}.
In the\highlight{ir approach}\remove{
setup proposed by \cite{abacha20233d}}, the query consists of an organ that may or may not contain a tumor. 
Successful retrieval requires accurately matching the tumor status, i.e., whether a tumor is present, and if present, correctly identifying its stage\remove{ if a tumor exists}. 
\remove{In the setup proposed by}\highlight{The experimental setup from} \cite{abacha20233d}, \highlight{relies on data sourced from four tasks} \remove{from}\highlight{of} the medical segmentation Decathlon (MSD) challenge dataset \remove{are used}\cite{antonelli2022medical}. The tumor segmentation is taken from the \remove{tumor masks published by the challenge organizers}\highlight{available ground truth label masks} \cite{antonelli2022medical}. The organ segmentation is performed using the TotalSegmentator segmentation model \cite{wasserthal2023totalsegmentator}. The combined information of organ segmentation and tumor segmentation is used to extract morphological information, e.g., size, number of lesions, location, and overlapping regions. \remove{This information is then used to create}\highlight{Finally, the} tumor stag\highlight{es}\remove{ing} based on \highlight{the} TNM staging standard \cite{sobin2009tnm} \highlight{are derived}.
The TNM staging \highlight{system basically relies on the following parameters:} T describes the size of the tumor and any spread of cancer into nearby tissue; N describes the spread of cancer to nearby lymph nodes; and M describes the metastasis (spread of cancer to other parts of the body). The setup proposed by \cite{abacha20233d} does not include lymph nodes and metastasis due to the unavailability of the related information for the MSD dataset. The staging information based on tumor size (i.e. T)  is used to create the train/test or database/query splits and the evaluation of retrieval approaches. 
The initial setup proposed by \cite{abacha20233d} is shown in \Cref{fig:overview-dataset}. 


The benchmark proposed by \cite{abacha20233d} relies on two key assumptions that limit its applicability to larger datasets. First, it assumes that \textbf{segmentation of each organ is available}, which requires either manual time-consuming delineation or an algorithmic segmentation solution, for example, TotalSegmentator \cite{wasserthal2023totalsegmentator}. While AI-based segmentation is a feasible option for \highlight{many} small \highlight{to medium-sized medical (volumetric) image} datasets, it becomes computationally expensive for larger datasets, which are more representative of real-world scenarios \highlight{where CBIR applications become relevant.}
Second, \cite{abacha20233d} created \textbf{separate datasets} for each organ. \remove{This means that}\highlight{For example}, the dataset (and consequently the search space) for colon tumor flagging and staging includes only slices of scans containing the colon. Similar setups are used for the liver and other organs, resulting in four separate datasets for the four organs. While th\remove{is}\highlight{e} benchmark \highlight{in \cite{abacha20233d}} demonstrates the potential of CBIR systems for tumor flagging and staging, its reliance on these assumptions makes it \remove{impractical}\highlight{difficult to apply to}\remove{for} larger datasets, thereby limiting a realistic evaluation of the algorithms. \highlight{Moreover, the presented test scenarios, which assume separated datasets for each organ are not viable in real-world scenarios where data from all organs (or anatomic regions) are stored in the same PACS. Thus, a reality-inspired test set with scans of all relevant anatomical regions would allow a more practical and extended evaluation.}
\remove{For example, in a clinical setting with huge amounts of imaging data, relying on separate datasets for each organ would hinder the ability to efficiently retrieve relevant cases, ultimately complicating the process and reducing the effectiveness of CBIR systems.}
Moreover, the criteria for data splits are not clearly defined. Despite these limitations, \remove{the benchmark}\highlight{\cite{abacha20233d}} provides a valuable starting point for assessing CBIR systems \highlight{in the context of tumor retrieval and staging}. To \highlight{further} enhance the evaluation process\remove{further}, implementing a more generalized dataset and an automatic selection of cases combined with randomization \remove{could}yield a more comprehensive assessment of the algorithms, ultimately improving their applicability in real-world clinical scenarios.

\subsubsection{Re-ranking}

\highlight{Building upon the foundation of CBIR systems for tumor retrieval, an additional challenge remains in optimizing the relevance of retrieved results. Information retrieval systems aim to provide users with the most relevant results for their queries according to a similarity score. However, initial retrieval results often require refinement to increase the relevance of retrieved information. This refinement process, known as re-ranking, has become an essential component in modern retrieval systems, particularly in CBIR \cite{mei2014multimedia}. 
Re-ranking refers to the process of modifying the order of initially retrieved results to better align with user preferences and requirements. Over the years, numerous approaches have emerged to address this challenge, employing diverse strategies that go beyond pairwise similarity measures \cite{mei2014multimedia, pedronette2014scalable, zhang2020understanding, zhong2017re, chum2007total}. 
One approach is relevance feedback, which involves collecting explicit or implicit input from users about the relevance of specific results. This feedback is then used to adjust the ranking, ensuring that more relevant items appear higher in subsequent searches \cite{pedronette2014scalable}.
Another approach involves learning-based algorithms, which utilize learning-based models to optimize ranking \cite{lee2022correlation, tan2021instance, shao2023global}. These algorithms analyze features extracted from the data, such as semantic content, visual characteristics, or user interaction patterns, to improve the ranking process. By training models on these features, the system can predict and adjust the relevance of search results, leading to more accurate and personalized retrieval outcomes \cite{zhang2020understanding}.

More recently, techniques that incorporate contextual information have gained prominence in re-ranking. One such method is ColBERT \cite{khattab2020colbert} (Contextualized Late Interaction over BERT \cite{devlin2018bert}). ColBERT addresses the limitations of traditional methods by encoding both documents and queries into rich, multi-vector representations. Instead of relying on single vector embeddings, ColBERT creates an embedding for each token in the query and document. Relevance is then measured by computing the total maximum similarities between each query vector and all vectors within the document. This late interaction architecture allows for a refined and contextually aware retrieval process \cite{khattab2020colbert}. Although ColBERT was originally developed for text retrieval, we propose to adopt its contextual late interaction principle for content-based 3D medical image retrieval.}

\subsection{Contribution}

\remove{In this study, we build upon the work of }\highlight{This study expands upon the work of} \cite{abacha20233d} by providing a more comprehensive \remove{assessment}\highlight{evaluation} of \remove{the algorithm}\highlight{a 3D medical CBIR system} on larger\highlight{, more realistic} datasets. \remove{To achieve this, we removed the two assumptions of pre-existing segmentations for database cases and separate databases for each organ. Additionally, we propose a sampling scheme to create databases that address the limitations of hand-picked cases, ensuring a more representative distribution of cases.
The contributions of this work are as follows:}
\highlight{We address limitations of prior work by removing the assumptions of pre-existing organ segmentations and organ-specific databases. Additionally, we introduce a novel sampling scheme to construct databases that better represent the true distribution of disease stages. Furthermore, we introduce an innovative re-ranking strategy that considers the 3D image context. The primary contributions of this work are:}

\begin{itemize}
    \item \remove{Developed organ-specific databases (colon, liver, lung, pancreas) with balanced stage distributions through systematic sampling}
    \highlight{\textbf{Organ-Specific Databases with Balanced Stage Distributions:} We propose a systematic sampling method to create four organ-specific databases (colon, liver, lung, pancreas) to ensure balanced representation of different tumor stages.}
    \item \remove{Introduced organ-agnostic database construction reflecting real-world PACS heterogeneity}
    \highlight{\textbf{Organ-Agnostic Database for Real-World Applicability:} We developed an organ-agnostic database to better reflect the heterogeneous nature of clinical PACS systems that allows for more realistic evaluation and deployment of CBIR systems.}
    \item  \remove{First adaptation of ColBERT's late interaction mechanism}\highlight{\textbf{ColBERT Adaptation for Volumetric CBIR and Segmentation-Free Retrieval:} We propose a novel adaptation of the ColBERT late interaction method, originally developed for text retrieval} \cite{khattab2020colbert}\remove{for volumetric CBIR, enabling context-aware re-ranking of 3D medical images and elimination of segmentation dependencies via implicit ROI localization}\highlight{, for volumetric CBIR that enables context-aware re-ranking of 3D medical images, and eliminates the requirement for pre-existing segmentations by implicitly localizing relevant Regions of Interest (ROIs).}
    \item \remove{Comprehensive quantitative evaluation across four tumor sites with statistical validation of results}
    \highlight{\textbf{Comprehensive Quantitative Evaluation with Statistical Validation:} We conducted a comprehensive quantitative evaluation of our approach across four distinct tumor sites, three feature extractors, and two re-ranking methods. }
\end{itemize}

\section{Material and Methods}\label{sec: material and methods}

\remove{In the following sections, we outline the methodologies employed in our study. First, in Section 2.1 we describe the vector database and indexing techniques used to facilitate image retrieval. Next, we detail the feature extractors utilized for generating embeddings from the imaging data in Section 2.2. Following this, we explain the configuration of the query and database setups, including both organ-specific and organ-agnostic setups in Section 2.4 and Section 2.5. In Section 2.6, we discuss the search and retrieval process based on slice information. Finally, we introduce our re-ranking method, which refines the retrieval results based on the information derived from the 3D volumes.}

\subsection{Vector Database and Indexing}
\label{sec: Inexing}

\remove{In image search, a database contains all image representations, known as embeddings, along with their associated metadata, such as annotations. 
Users or systems can make queries to find specific images through methods like uploading a reference image or providing a text description. 
The objective is to find images in the database that are similar to the query. 
In this study, the search process involves comparing a query image (containing a tumor or not) to those in the database to find the most similar one based on the cosine similarity of their embeddings. 
In the current setup, we do not utilize any metadata in this process. 
While similarity searches are straightforward with small datasets, they become more complex as the database size increases.}

\highlight{In CBIR, search involves comparing query images against a database of image representations, also referred to as embeddings, to find similarities. In this study, we use cosine similarity to compare embeddings of query images (containing a tumor or not) without using metadata of any kind.}
I\remove{n content-based image search, i}ndexing refers to establishing a structure for the efficient storage and retrieval of embeddings. \remove{The embeddings are created based on only pixel data, i.e., visual character of the images.  
There are multiple indexing methods to choose from.}\highlight{Based on} \remove{according to}the findings in \cite{jush2024medical, 10966872}, \highlight{we selected Hierarchical Navigable Small World (}HNSW\highlight{)} \cite{malkov2018efficient}\remove{was selected} as index\remove{for this study}.\remove{Various indexing solutions exist for storing and searching vectors, and in this research, we utilized} The Facebook AI Similarity Search (FAISS) package \highlight{is used for implementation,}\remove{, which facilitates rapid similarity searches }\remove{In our approach, the embeddings are indexed in a vector database using}\highlight{ specifically, the} HNSWFlat \highlight{index} \cite{johnson2019billion}. 
\remove{The overall process involves using feature extractors on slices of each volumetric image for those slices These embeddings are then stored in a structured format known as a vector database. The stored embeddings are organized into a search index, which facilitates efficient similarity-based retrieval of images when a query is made.}\highlight{The overall process can be summarized as extracting embeddings from slices of volumetric images and storing them in a searchable vector database for efficient similarity-based retrieval.}

\subsection{Feature Extractors}
\label{sec:feature-extractors}

We used three pretrained models as feature extractors\remove{.}\highlight{, selected to represent a diverse range of training strategies and architectural approaches. Specifically, we included a model that leverages an ensemble of self-supervised and contrastively trained components trained on natural images (DreamSim trained on ImageNet \cite{5206848}), a model trained with supervised learning on a large medical image dataset (SwinTransformer \cite{liu2021swin} on RadImageNet \cite{mei2022radimagenet}), and a model trained with contrastive learning using paired medical images and text (BioMedClip \cite{zhang2023biomedclip}).}
Previous studies have demonstrated the \highlight{efficacy of}  \remove{feasibility of using} pretrained self-supervised models based on the DINO framework \cite{caron2021emerging, oquab2023dinov2, fu2023dreamsim} for medical retrieval tasks \cite{denner2024leveraging, jush2024medical}. 
In this work, we used DreamSim \remove{as it is an ensemble model}\cite{fu2023dreamsim}\highlight{ as a representative of this class, specifically we used the ensemble version of DreamSim that consists of DINO model plus CLIP \cite{radford2021learning} and OpenCLIP \cite{cherti2023reproducible} and therefore includes strengths of both self-supervised visual representation learning (DINO) and contrastive image-text learning (CLIP/OpenCLIP)}. Additionally, a SwinTransformer \cite{liu2021swin} trained on RadImageNet \cite{mei2022radimagenet} is included based on \remove{the comparable performance to the self-supervised models reported by}\highlight{its reported competitive performance compared to self-supervised models in medical image retrieval} \cite{jush2024medical, 10966872}\highlight{, offering a strong baseline trained directly on a large-scale medical dataset}. 
Furthermore, the BioMedClip model, \remove{which has been employed}\highlight{previously used} for tumor retrieval in \cite{abacha20233d}, was incorporated \remove{for}\highlight{to provide a point of} comparison \highlight{to existing work in the field and to assess the transferability of a model trained with multi-modal (image and text) data}. 
\highlight{While fine-tuning pre-trained models on task-specific data can potentially yield further performance gains, we focused on evaluating the zero-shot transfer capabilities of these pre-trained models in this study. This allows us to assess their inherent ability to extract relevant features without task-specific or modality-specific model adaptations.} The choice to exclude convolution-based models in this study was mainly motivated by recent advancements of transformer-based models and the DINO framework. It should be noted that this study does not \highlight{aim at exhaustively benchmarking all available image embedding models in the context of 3D CBIR. Still we believe that the selected models are diverse enough to provide valuable insights on the capabilities of modern vision embeddings in this context.}

\subsection{Dataset}

\remove{In line with}\highlight{Following} \cite{abacha20233d}, \remove{our study employed four organs - liver, lung, pancreas, and colon - sourced from the}\highlight{we utilized publicly available data from the} MSD challenge \cite{antonelli2022medical}\highlight{, specifically, the data from task 3 (colon tumor segmentation),  task 6 (liver tumor segmentation), task 7 (lung tumor segmentation) and task 10 (pancreas tumor segmentation)}. 
The volumes chosen for the query set and database originate from the MSD training set. 
\remove{An overview of the number of volumes and slices is presented in}\highlight{The aggregated dataset contains overall 601 3D volumes with 115,899 2D slices, as detailed in} \Cref{tab:msd overview}. \remove{The dataset consists of 601 3D volumes and 115899 2D slices.}This data is utilized in the construction of the query and database sets for \remove{the}\highlight{our} experiments.\remove{It is assumed that segmentation masks are available for the query sets.}\remove{The creation of databases consists of three phases. Initially, we generated organ-specific databases that were limited to slices of the examined organ, following the methodology outlined in \cite{abacha20233d}. 
Next, we expanded our approach to include all slices from the relevant cases, enhancing the database's comprehensiveness. Finally, we developed an organ-agnostic database that integrates data from multiple organs, facilitating a unified analysis.} \remove{It is important to note that organ and tumor segmentation masks were essential for conducting the experiments and evaluations. In this study,}\highlight{Tumor segmentation masks are taken from the MSD ground truth masks \cite{antonelli2022medical} and organ} segmentation masks were created for all the 3D volumes utilizing the TotalSegmentator model \cite{wasserthal2023totalsegmentator} to facilitate comprehensive comparisons.\remove{Although the purpose of the organ-agnostic database is to circumvent the need for segmenting every case, these segmentations were initially required to support our analysis.}
\Cref{fig:overview-dataset} provides an overview of this process. 

\begin{table}[h]
\captionsetup{width=\textwidth} 
    \caption{\remove{Overview of MSD Training Set. Number of volumes and slices for}\highlight{Composition of the MSD challenge dataset, showing number of volumetric scans and axial slices per defined tasks; }\highlight{task 3:}\remove{the} colon \highlight{tumor segmentation}, \highlight{task 6:} liver \highlight{tumor segmentation}, \highlight{task 7:} lung \highlight{tumor segmentation}, and \highlight{task 10:} pancreas \highlight{tumor segmentation}\remove{task}.}
    \begin{tabular}{c|cc} 
        {MSD Tasks \remove{Organ}} & {\highlight{3D} Volumes} & {Slices} \\
        \hline
        \highlight{Task 3}\remove{Colon} & 126 & 13360 \\
        \highlight{Task 6}\remove{Liver} & 131 & 58507 \\
        \highlight{Task 7}\remove{Lung} & 63 & 17594 \\
        \highlight{Task 10}\remove{Pancreas} & 281 & 26438 \\
        \hline
        Total & 601 & 115899 \\
    \end{tabular}
    \label{tab:msd overview}
\end{table}

\begin{figure}[!htbp]
  \centering
  \includegraphics[trim=0.5cm 0cm 4cm 0.2cm, clip,width=\textwidth]{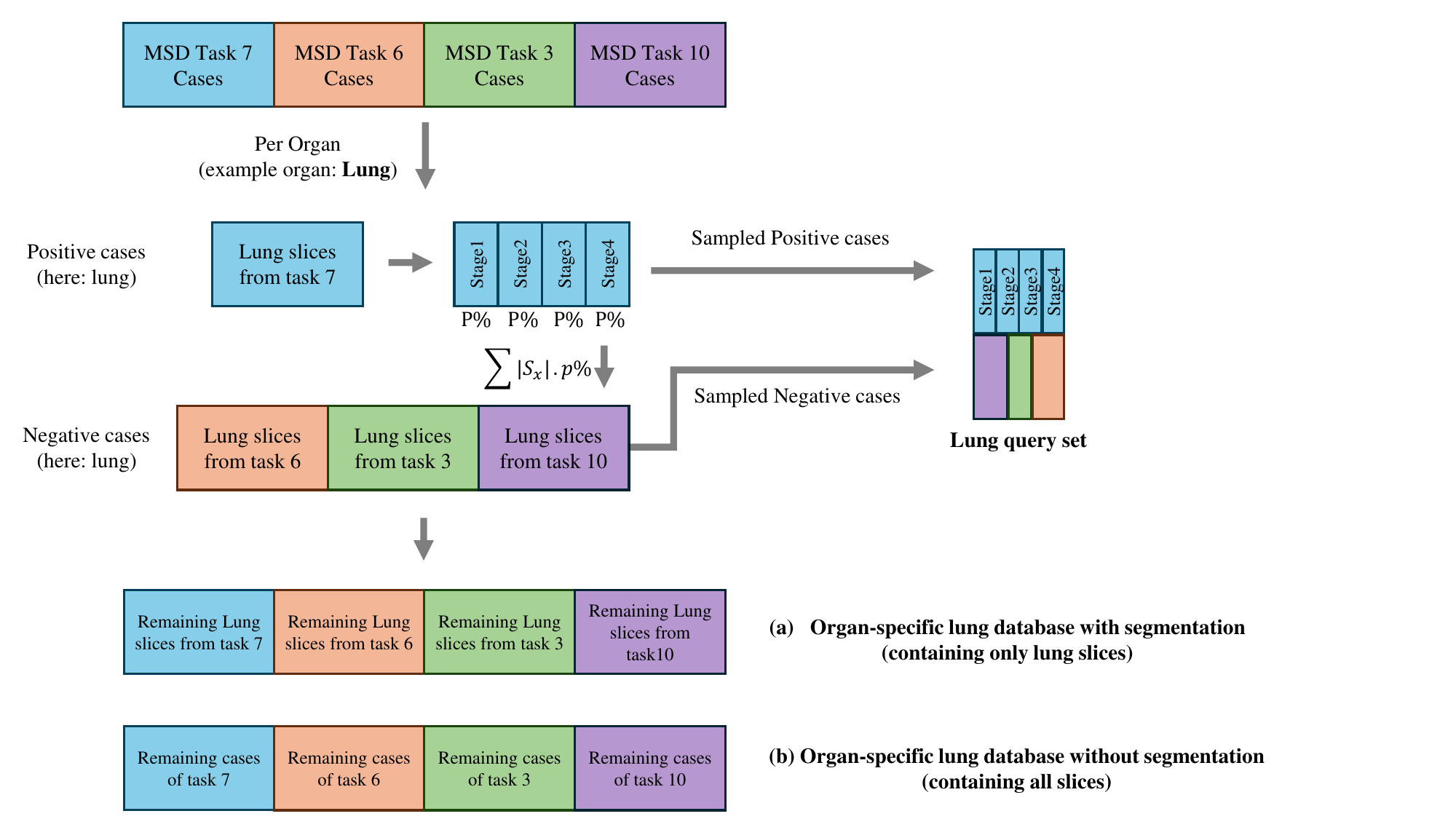}
  \caption{An overview of creating the organ-specific query and databases. For each \remove{task}\highlight{organ}, separate database and query sets are created. In (a), the segmentation masks are used to filter the slices containing lung, which limits the search space to specific lung regions. In contrast, (b) includes all slices in the search space.}
  \label{fig:overview-organ-specific}
\end{figure}

\subsection{Query Setup}
\label{sec: query}

\remove{We present two approaches for query setup: organ-specific and organ-agnostic.}
\highlight{We create two different query datasets for our experiment: an organ-specific setup and organ-agnostic setup.}

\subsubsection{Organ-specific}
\label{sec: query organ-specific}
\remove{Our organ-specific approach begins with segmenting all 3D volumes from the four MSD tasks using the TotalSegmentator model \cite{wasserthal2023totalsegmentator}. For each organ-specific database, we construct the query data by combining positive and negative sets.}
\highlight{The organ-specific query set combines tumor-positive and tumor-negative cases, sampled across the selected MSD tasks.}
\remove{The positive slice sets comprise slices containing tumors, categorized into different staging groups denoted as ($S_x$). We randomly sample ($p\%$) from each staging group, with ($p$) set to $25\%$ here. As shown in Figure 2, for a specific organ like the lung, the positive cases are derived from lung slices of the lung MSD task.
The negative slice sets consist of slices from the three remaining MSD tasks that contain the organ under examination. For instance, when building a lung query set, the negative cases are sourced from lung slices of the pancreas, liver, and colon MSD tasks. The sum of positive cases is calculated as $T_p = \sum S_x.p$, where $T_p$  represents the total number of positive cases. We then sample an equal number of cases from the negative slice sets.}
\highlight{For each organ (e.g., lung from Task 7), we created positive and negative query sets. Positive cases were defined as $p\%$ (here 25\%) of tumor-containing cases per stage ($S_1, .., S_4$), resulting in $T_p = \sum_{x=1,..,4} |S_x| \cdot p$ total positive cases. Negative cases were matched to the number of positive cases, and consisted of non-tumor slices of the same organ, but taken from other tasks (e.g., slices that contain lung from Tasks 3, 6, and 10 scans).}
\highlight{We repeated this sampling process 10 times with different random seeds (sampling is performed with replacement), generating distinct query/database splits for statistical reliability. Embedding counts and case distributions are detailed in} \Cref{tab:organ-specific table} \highlight{(Query Vol. and Query Emb. columns)}. \highlight{To address the potential correlation between slices within a single 3D volume, we ensured that all slices from a given volume were kept together within the same query/database split. This was achieved by splitting the data at the volume level, rather than the slice level.} \Cref{fig:overview-organ-specific}\highlight{ visualizes the lung-specific query dataset generation as an example.}

\remove{We conduct the experiment for every organ 10 times to achieve representative sampling and statistical validity. This is done by generating separate query and database sets through the use of 10 random seeds for case sampling. The number of sample cases for each organ, along with the corresponding number of embeddings, is detailed in Table 2.}

\begin{table}[h]
    \captionsetup{width=\textwidth}
    \caption{Overview of number of volumes (Vol.) and embeddings (Emb.) for organ-specific query and database. The notation $P.+N.$  indicates the inclusion of both positive and negative cases. The query set remains consistent across database configurations with (w.) and without (wo.) segmentation. The symbol $\pm$ shows the standard deviation of slice counts across 10 experiments using 10 seeds.}
    \scriptsize
    \begin{tabular}{c|cc|cccc} 
        \multirow{2}{*}{{Organ}} &  \multicolumn{1}{c}{{Query Vol.}} & \multicolumn{1}{c|}{{Query Emb.}} & \multicolumn{1}{c}{{Database Vol.}} & \multicolumn{1}{c}{Database Emb.} & \multicolumn{1}{c}{Database Emb. }   \\
        & (P.+N.) & {(P.+N.)} & (P.+N.) & w. Segmentation & wo. Segmentation \\ 
        \midrule
        Colon & 60 & $5618 \pm 562$ & 535 & $6647 \pm 562$  & $104731 \pm 1283$  \\
        Liver & 62 & $6373 \pm 344$ & 533 & $40614 \pm  344$  & $97802  \pm 1247$ \\
        Lung & 28 & $4119 \pm 445$ & 565 & $43202 \pm  445$  & $108709 \pm 1042$\\
        Pancreas & 138 & $5378 \pm 326$  & 451 & $18016 \pm 324$ & $88879 \pm  1865$ \\

    \end{tabular}
\label{tab:organ-specific table}
\end{table}

\subsubsection{Organ-agnostic}
\label{sec: query organ-agnostic}

\begin{figure}[!htbp]
  \centering
  \includegraphics[trim=0.5cm 0.5cm 0cm 0.2cm, clip,width=\textwidth]{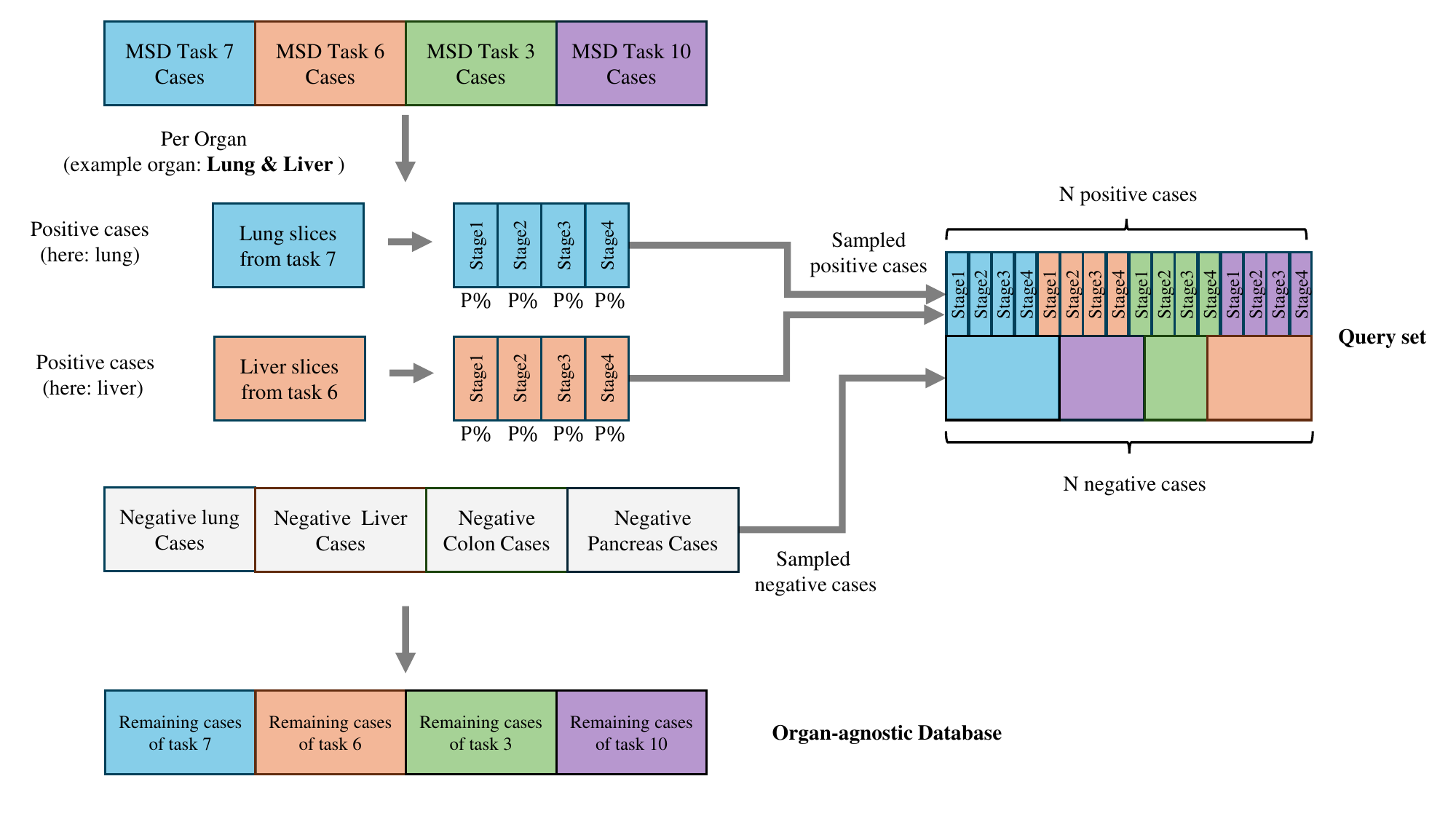}
  \caption{An overview of creating the organ-agnostic query \highlight{set} and database. The query sets and the database encompass all organs and tumor stages. Negative cases are also present in both the query set and the database.}
  \label{fig:overview-unified-database}
\end{figure}

\remove{Here, query set contains cases from four organs. 
$p\%$ of the dataset for each organ is selected as positive cases, and an equal number of negative cases is sampled accordingly.}
\highlight{The organ-agnostic query set was created by including cases from all four organs (MSD tasks 3, 6, 7, and 10). For each organ, we sampled $p\%$ (here $25\%$ of the tumor-containing cases as positive cases and sampled an equal number of non-tumor cases to maintain a balanced query set.} \Cref{fig:overview-unified-database} provides \highlight{an overview of the data generation} and \Cref{tab:organ-agnostic database} (Query Vol. and Query Emb. columns) shows the detailed number of cases and slices. \highlight{The organ-agnostic set was also created by splitting the data at the volume level, ensuring that all slices from a single volume were included in either the query set or the database.}

\begin{table}[!tbph]
    \centering
    
    \caption{Overview of number of volumes (Vol.) and embeddings (Emb.) for organ-agnostic query and database. The notation $P.+N.$  indicates the inclusion of both positive and negative cases. The symbol $\pm$ shows the standard deviation of slice counts across 10 experiments using 10 seeds.}
    \scriptsize
    \begin{tabular}{c|cc|cc} 
        \multirow{2}{*}{{Organ}} & \multicolumn{1}{c}{{Query Vol.}} & \multicolumn{1}{c|}{{Query Emb.}} & \multicolumn{1}{c}{{Database Vol.}} & \multicolumn{1}{c}{{Database Emb.}} \\
        & (P.+N.) & (P.+N.) & (P.+N.) & \remove{(wo. Segmentation)}\highlight{ (P.+N.) } \\ 
        \midrule
        All organs & $244 \pm 4$ & $19920 \pm 761$ & $356  \pm 4$ & $65377 \pm 2224$ \\ 
    \end{tabular}
    \label{tab:organ-agnostic database}
\end{table}

\subsection{Database Setup}
\label{sec: database}

\remove{We present three approaches for database setup: organ-specific with segmentation, organ-specific without segmentation, and organ-agnostic.}
\highlight{Our experiments comprise the following different database setups: organ-specific with segmentation, organ-specific without segmentation, and organ-agnostic.}

\subsubsection{Organ-specific with Segmentation}
\label{sec: organ-specific with segmentation database}

\remove{In this approach,} We create\highlight{d} four separate databases, each containing positive and negative cases\remove{slices from each organ set}. \remove{Subsequent to}\highlight{After} forming the query set \highlight{(}as detailed in \Cref{sec: query organ-specific}\highlight{)}, the \remove{database is constituted by the} remaining $75\%$ of cases \highlight{constituted the database}.  \remove{An example overview is shown} 
\highlight{For example, as illustrated} in \Cref{fig:overview-organ-specific} (a) \highlight{for the lung, the search space is restricted to only the lung slices (with or without tumor).}\remove{with lung as the example organ. By presuming the availability of segmentations, we restrict the search space to only lung slices.}
\remove{The d}\highlight{D}etails of the number of cases and embeddings can be found in \Cref{tab:organ-specific table} \remove{,}\highlight{(}"Database Vol." and "Database Emb. w. Segmentation"\highlight{columns)}.

\subsubsection{Organ-specific without Segmentation}

\remove{Creating segmentation masks, even when automated, for each instance in the database can be computationally intensive and impractical for real-world applications.
Consequently, we consider the cases included in the database outlined in}
\highlight{Here, we used the same cases as in the "with segmentation" approach (}\Cref{sec: organ-specific with segmentation database}\highlight{)} and remove\highlight{d} the assumption that \highlight{organ} segmentation \highlight{mask}s are available. 
\remove{We perform experiments with the complete set of slices to evaluate the performance of embeddings in localizing organ regions.}\highlight{As a result, the search space includes all slices, as the example shown}
in \Cref{fig:overview-organ-specific} (b). \remove{the search space includes all slices.}\remove{The d}\highlight{D}etails of the number of cases and embeddings can be found in \Cref{tab:organ-specific table}\remove{,}\highlight{ (}\remove{columns}"Database Vol." and "Database Emb. wo. Segmentation"\highlight{columns)}. 
As these databases encompass all the slices, they are 1.5 to 4.5 times larger than the databases \highlight{described in} \Cref{sec: organ-specific with segmentation database}. \remove{The potential benefits of prior segmentation are analyzed in Section 3.}

\subsubsection{Organ-agnostic}

\remove{In a more realistic scenario, all data is stored in a single database. We propose a setup where slices from all tasks are combined into one comprehensive database,}\highlight{To simulate a more realistic scenario where all data is stored in a single database, we created an organ-agnostic database by combining images from all tasks,} as \highlight{the example }shown in
\Cref{fig:overview-unified-database}. 
\remove{In our experimental design, once the query set is established}\highlight{After establishing the query set} (described in \Cref{sec: query organ-agnostic}), the \remove{rest of the}\highlight{remaining} cases \remove{constitute the}\highlight{are stored in a single, unified} database. 
\remove{Consequently, only one database is created, which is ideal in practical scenarios.}\highlight{Here we do not make use of any information derived from image segmentation masks.} \Cref{tab:organ-agnostic database} shows the detailed number of cases and slices \highlight{("Database Vol." and "Database Emb." columns)}.

\subsection{Search and Retrieval}
\label{sec: search and retrieval}

The search is conducted \remove{using}\highlight{by comparing the similarity} of embeddings obtained from the slices of the image volumes. 
The most straightforward retrieval method involves \remove{comparing}\highlight{retrieving for} a 2D query slice $q$ with the most similar 2D slice \highlight{$s^*$ from} the database\remove{$s^*$}. This is done by identifying the slice embedding that maximizes the cosine similarity with the embedding linked to $q$:
\begin{equation}\label{eq:similarity_search}
    s^* 
    = \underset{s \in \text{Database}}{\mathrm{argmax}} \frac{\langle\phi(s),\phi(q)\rangle}{\Vert \phi(s) \Vert_2 \Vert \phi(q) \Vert_2}
    = \underset{s \in \text{Database}}{\mathrm{argmax}} \left\langle v_s,\frac{\phi(q)}{\Vert \phi(q) \Vert_2}\right\rangle
\end{equation}
where $\langle\cdot,\cdot\rangle$ denotes standard scalar product, $\Vert\cdot\Vert_2$ the euclidean norm, $\phi$ the embedding mapping and $v_s = \phi(s)/\Vert \phi(s) \Vert_2$ the pre-computed, normalized embedding associated to slice $s$ stored in \remove{a vector index}\highlight{the vector database}.
Given the query volume $V_Q = [q_1, ..., q_n]$, the system retrieves the most similar slice $s_i^*$ from the database for each slice $q_i$ in the query volume, $V_Q$, using \eqref{eq:similarity_search}. The associated volume ID and its similarity score are then recorded in a hit\highlight{-}table. 
\remove{Next, w}\highlight{W}e implement the Count-base aggregation method from \cite{jush2024medical}, which utilizes \remove{the}\highlight{a} hit-table to determine the volume $V_R$ that has the highest number of hits for the given query volume.
To ensure comparability with \cite{abacha20233d}, for each slice query, the 20 most similar slices are considered and \remove{eventually}the top-k similar volumes are retrieved per each query volume. 
Moreover, \highlight{based on} the hit-table\remove{calculates} the maximum similarity score (Max-Score) and the total similarity score (Sum-Sim) \highlight{are calculated}, and two \highlight{additional} top-k volume sets \cite{abacha20233d} \highlight{are obtained}. The computation of Max-Score and Sum-Sim follows \cite{abacha20233d} \highlight{equation 2 and 3, respectively}.

\subsection{Re-ranking}

\subsubsection{C-MIR: Colbert-inspired Medical Image Retrieval and Re-ranking}\label{sec:cmir_method}

\remove{In the field of information retrieval, re-ranking refers to the process of modifying the order of retrieved results to better align with the user's preferences and requirements. This adjustment can be achieved using various techniques, such as relevance feedback, where user input is utilized to refine results, or learning-based algorithms that leverage relevant features to optimize ranking.
Additionally, contextual information, such as user behavior and temporal relevance, can enhance the accuracy of rankings. 
A notable method that incorporates contextual information is ColBERT (Contextualized Late Interaction over BERT), cf. \cite{devlin2018bert} \cite{khattab2020colbert}. 
In ColBERT, documents and queries are encoded into multi-vector representations, and relevance is measured through interactions between these sets of vectors. 
Per each token in the query and document, ColBERT creates an embedding and measures relevance by computing the total maximum similarities between each query vector and all vectors within the document, allowing a refined and contextually aware retrieval process \cite{khattab2020colbert}.}

\begin{figure}[!t]
  \centering
  \includegraphics[trim=0cm 0cm 0cm 0cm, clip,width=\textwidth]{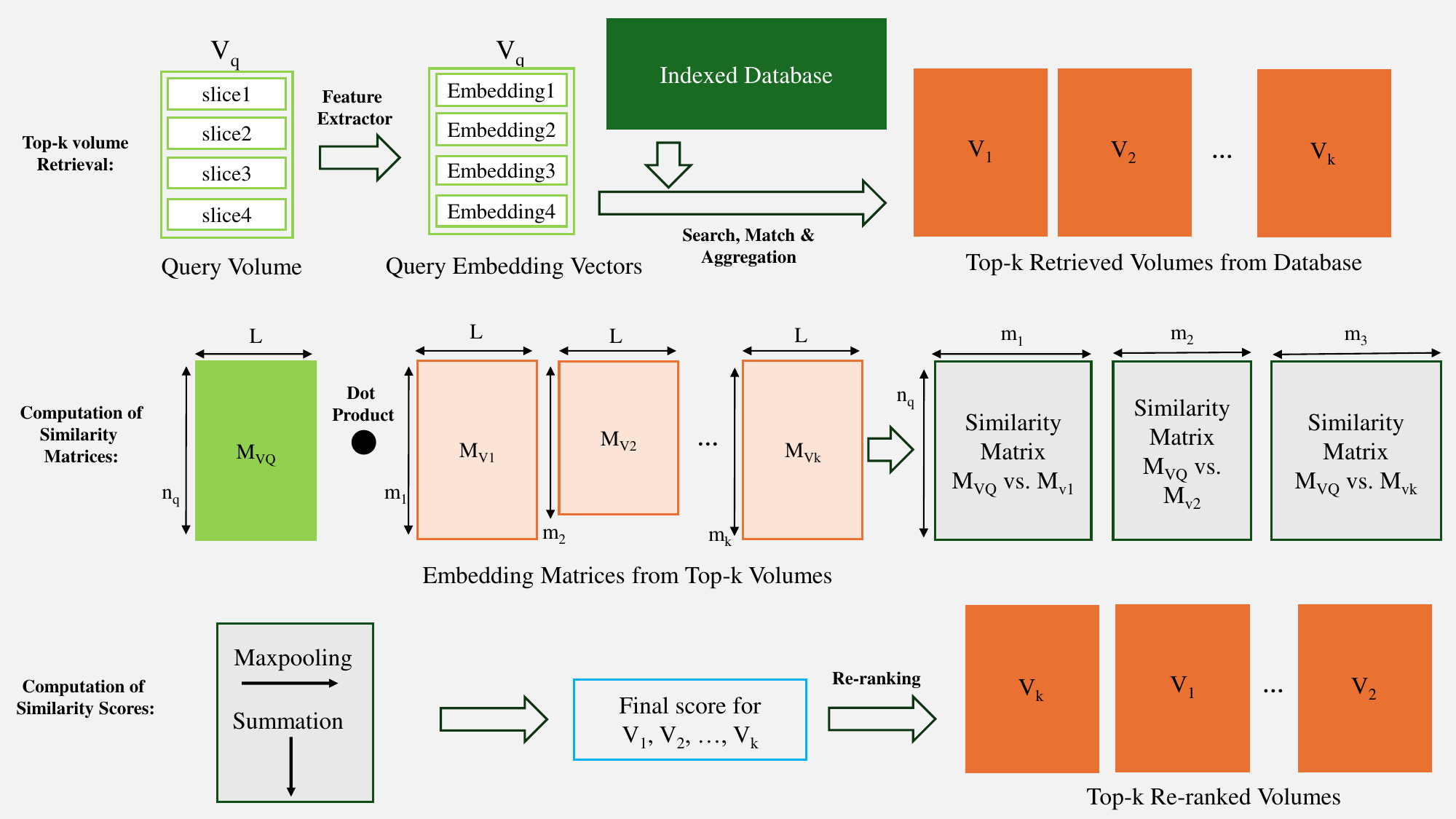}
  \caption{Overview of C-MIR. Image slices vector embeddings are created as explained in \Cref{sec:feature-extractors}. The top-k volumes are retrieved based on aggregation criteria presented in \Cref{sec: search and retrieval}. The embedding matrices for the query and the top-k retrieved volumes are utilized to calculate similarity matrices. The rows and columns of all similarity matrices are pooled and summed to compute a rank score per volume, see \Cref{sec:cmir_method}. Finally, the retrieved volumes are re-ranked based on their rank score.}
  \label{fig:re-ranking}
\end{figure}

Inspired by ColBERT \highlight{\cite{khattab2020colbert}}, here we propose a re-ranking method. 
To create an analogy to the ColBERT method, each slice can be \remove{mapped to}\highlight{interpreted as} a word, and each volume can be \remove{mapped to}\highlight{interpreted as} a passage. 
Instead of the BERT encoder \highlight{\cite{devlin2018bert},} for the image retrieval task, the pre-trained vision models are used (see \Cref{sec:feature-extractors}). 
A brief overview of the method is shown in \Cref{fig:re-ranking}. We call this re-ranking method ColBERT-Inspired Medical Image Retrieval and Re-Ranking (C-MIR).  

Once the top-k volumes are selected according to the similarity of individual vectors and the aggregation criteria described in \Cref{sec: search and retrieval}, the selected volumes undergo a re-ranking process:

\paragraph{Step 1: Constructing the Embedding Matrix} For any volume $V=[v_1,...,v_n]$ with $n$ slices $v_1,...,v_n$ we can compute the embedding matrix $M_V$ of dimension  $n \times L$: 
\begin{equation}\label{eq:M_v}
    M_V = 
    \left[
    \frac{\phi(v_1)}{\Vert\phi(v_1)\Vert_2}, 
    ...,
    \frac{\phi(v_n)}{\Vert\phi(v_n)\Vert_2} 
    \right]
\end{equation}
where $\phi$ assigns each slice to a vector of a constant length $L$, known as an embedding vector. Thus, $M_V$ consists of a collection of embedding vectors derived from the slices.

\paragraph{Step 2: Embedding Matrix Similarity} Assuming another volume $W=[w_1,...,w_m]$ of size $m \times L$ we can compute the similarity matrix of size $n\times m$ via:
\begin{equation}\label{eq:similarity_matrix}
    \mathrm{SIM}(M_V, M_W)
    = \left[\frac{\langle \phi(v_i),\phi(w_j) \rangle}{\Vert\phi(v_i)\Vert_2 ~ \Vert\phi(w_j)\Vert_2} \right]_{\substack{i=1,...,n\\j=1,...,m}}
\end{equation}
The entry $(i,j)$ of this matrix contains the cosine-similarity score of the embeddings related to slice $i$ of volume $V$ and slice $j$ of volume $W$ (the extracted embeddings undergo $L_2$ normalization in a postprocessing step; consequently, the dot product becomes equivalent to the cosine similarity).
Considering $V_Q$ as the query volume, our re-ranking process begins with calculating the embedding matrix $M_{V_Q}$ according to \eqref{eq:M_v}. Next, we determine the embedding matrices $M_{V_1}, \ldots, M_{V_M}$ for each unique retrieved volume $V_1, \ldots, V_M$ from the initial search, with the top $M=20$ considered (refer to \Cref{sec: search and retrieval}). Subsequently, we calculate the similarity matrices $\mathrm{SIM}(M_{V_Q}, M_{V_k})$ for $k = 1, \ldots, M$. In the following step, the similarity matrices are assessed by establishing a final rank score, which is used to reorder the volumes.

\paragraph{Step 3: Computing Final Rank Scores and Re-ranking}

To calculate the final rank score for each volume $V_k$, we first apply max\highlight{-}pooling row-wise to the similarity matrix $\mathrm{SIM}(M_{V_Q}, M_{V_k})$. This process identifies the slice in $V_k$ that has the highest cosine similarity to a specific slice in $V_Q$. The resulting vector, which has a length of $n$, is then summed up in order to derive the overall maximum slice similarity, which serves as the final rank score (RS). For $k=1,...,M$, we perform:
\begin{equation}\label{eq:RS-score}
    \mathrm{RS}(V_k)
    = \sum_{i=1}^n \max_{j=1,...,m_k} \mathrm{SIM}(M_{V_Q}, M_{V_k})_{i,j}
\end{equation}
where ${V_k}_j$ represents the $j$-th slice of volume $V_k$ and $m_k$ indicates the total number of slices in ${V_k}$. $n$ is the number of slices in $V_Q$.
The top-k volumes are then re-ranked based on their rank scores (RS), meaning that the volume with the highest score is the most relevant volume, and the volume with the lowest score is the least relevant volume in the top-k results, considering the whole volume slices.

\subsubsection{Reciprocal Rank Fusion}

\highlight{We compare our re-ranking approach against single aggregation modes and Reciprocal Rank Fusion (RRF) re-ranking approach \cite{bruch2023analysis, cormack2009reciprocal}. RRF is a meta-ranking technique that combines multiple retrieval lists obtained from different retrieval methods in order to leverage the complementary strengths of those \cite{jin2024matching, rackauckas2024rag}. In our setup, we use for RRF the three retrieval methods, i.e., Count-base, Sum-Sim, and Max-Score. Hence, given a query $V_Q$ we generate three ranked lists $L_1, L_2, L_3$ each containing the top 20 retrieved volumes for each method, i.e. $L_\ell = [V_{\ell, 1}, V_{\ell, 2}, ...,V_{\ell, 20}]$. For any volume $V \in L_1  \cup L_2\cup L_3$ the RRF score is then computed as}
\begin{equation} RRF(V) = \sum_{\ell=1}^{3} \frac{1}{k + \text{rank}(V, L_\ell)} \end{equation}
\highlight{where $\text{rank}(V, L_\ell)$ denotes $V$'s rank in the list  $L_\ell$. In case $V$ is not contained in $L_\ell$, the rank is set to $+\infty$, leading to a zero summand in the overall score.
The smoothing parameter $k$ is set to 60 following \cite{cormack2009reciprocal, chen2022out}. The final re-ranking is based on the RRF scores, i.e. the retrieved volume with the hightest RRF score is assigned the highest final rank.}

\section{Results and Evaluation}
\label{sec:results and evaluation}

This section contains a detailed quantitative evaluation of the retrieval results.
Additionally, Appendix \ref{secA1} includes a visual representation of \highlight{selected} retrieval outcomes for four cases, illustrating both failed and successful instances of tumor flagging and staging through CBIR, which serves to provide the reader with a conceptual framework to better \remove{understand}\highlight{contextualize} the quantitative results presented in this section.
In line with \cite{abacha20233d}, we used two metrics to evaluate the results: Precision at $k$ (P@k) and Average precision (AP). Precision at $k$ (P@k) is defined as 
\begin{equation}
    P@k = \frac{|{ \text{retrieved cases including tumor in top } k }|}{k} 
\end{equation}
where $k= 3, 5, 10$. 
\highlight{When evaluating information retrieval systems, precision and recall offer a general overview based on the top-k set of retrieved information. However, in many applications, the order in which documents are returned is crucial. Average Precision (AP) is a metric designed to capture this aspect, providing a single-value summary of ranking quality.}
\highlight{The} average precision (AP) \highlight{metric} is \remove{calculated using the formula:}\highlight{defined as:}
\begin{equation}
    AP = \sum_{n=1}^{10} ({R_n - R_{n-1}}){P_n}
\end{equation}
where $R_n$ and $P_n$ represent the Recall and Precision at the $n$-th \remove{rank}position \highlight{in the ranked list of top 10 retrieved cases \cite{kishida2005property, zuva2012evaluation, abacha20233d}}\remove{where $n=1,...,10$ (top 10 retrieved cases)}.
\highlight{The formula calculates a weighted average of the precisions at each rank, where the weights are the changes in recall between consecutive ranks. This measures how well the system ranks relevant information higher in the list. The results presented in the following sections show the means of the AP metric, computed across 10 repetitions of the entire experiment, each using a different random seed for case sampling.}
\remove{The results in the following sections show the averages of the metrics from 10 different experiments using 10 seeds for case sampling.}

\subsection{Tumor Flagging}
\label{sec: tumor flagging}

\remove{The following sections provide the results of the tumor flagging for organ-specific and organ-agnostic databases for the colon, liver, lung, and pancreas using BioMedClip, DreamSim, and SwinTransformer models as feature extractors. 
Three aggregation methods, as outlined in Section 2.6, are employed and evaluated alongside C-MIR to obtain top-k results.}

\subsubsection{Organ-specific Database} 
\Cref{tab:flagging combined w and wo} shows the performance of \highlight{re-ranking methods, i.e.,} C-MIR \highlight{and RRF} for tumor flagging in comparison with \remove{different}\highlight{the three vanilla} aggregation methods for four organs, three \highlight{different} feature extractors \remove{for}\highlight{and} the organ-specific databases with and without segmentation. 
The results of C-MIR are consistent regardless of the use of segmentation \highlight{mask}s, showing the \remove{method's}capacity \highlight{of the method} to localize the relevant regions effectively.
\highlight{In our evaluation,} C-MIR \remove{improves the results} is \highlight{consistently} the best-performing method for colon tumor flagging for all the models. 
C-MIR enhances the results for BioMedClip and SwinTransformer embeddings in liver tumor flagging \highlight{for the databases with segmentation}, but performance declines \remove{for}when utilizing DreamSim \highlight{embeddings}.
C-MIR \remove{enhances outcomes for}\highlight{is the best-performing method for} all models in lung tumor flagging.
For flagging pancreatic tumors, C-MIR slightly enhances results with BioMedClip but shows reduced performance for DreamSim and SwinTransformer \highlight{embeddings}.

\remove{DreamSim achieves the highest AP of $0.807$ for colon tumor flagging using C-MIR.} 
\highlight{C-MIR achieves the highest AP of $0.807$ for colon tumor flagging using DreamSim embeddings.} For liver tumors, the highest AP is $0.811$ with the Sum-Sim method, \remove{also }utilizing DreamSim \highlight{embeddings}. In lung tumor flagging, C-MIR \remove{with}\highlight{using} DreamSim \highlight{embeddings} stands out with an AP of $0.942$. Regarding pancreas tumor flagging, \remove{the DreamSim model leads in the count-base method}\highlight{count-base method using DreamSim embeddings lead}, achieving an AP of $0.802$ without segmentation and $0.797$ with segmentation.\remove{Overall, DreamSim surpasses other embeddings in this configuration.}
\highlight{The reported metrics represent the average values obtained from 10 experiments, each employing a different random seed for case sampling. A statistical analysis related these results is provided in} \Cref{sec: stastical analysis}

\renewcommand{\arraystretch}{1.3}
\begin{table}[!htbp]
  \centering
  \aboverulesep = 0pt
    \belowrulesep = 0pt
  \tiny
  \caption{Overview of \textbf{tumor flagging} results using \textbf{organ-specific databases} with and without segmentations.  \highlight{Reported metrics represent the average values across 10 experiments, each employing a different random seed for case sampling.} The bold-faced value in each sub-column shows the best method for each model. }
    \begin{tabular}{c|c|c|cccc||cccc}
          & \multirow{2}[0]{*}{Model} & \multirow{2}[0]{*}{Method} & \multicolumn{4}{c}{With Segmentation} & \multicolumn{4}{c}{Without Segmentation} \\
    \multirow{13}[0]{*}{\begin{sideways}Colon\end{sideways}} &       &       & \textcolor[rgb]{ .275,  .471,  .525}{p@3} & \textcolor[rgb]{ .275,  .471,  .525}{p@5} & \textcolor[rgb]{ .275,  .471,  .525}{p@10} & AP    & \textcolor[rgb]{ .275,  .471,  .525}{p@3} & \textcolor[rgb]{ .275,  .471,  .525}{p@5} & \textcolor[rgb]{ .275,  .471,  .525}{p@10} & AP \\ \hline
          & \multirow{4}[0]{*}{\begin{sideways}BioMedClip\end{sideways}} & C-MIR & \textbf{0.660} & \textbf{0.655} & \textbf{0.651} & \textbf{0.723} & \textbf{0.660} & \textbf{0.655} & \textbf{0.651} & \textbf{0.723} \\
          &       & \highlight{RRF} & \highlight{0.617} & \highlight{0.625} &	\highlight{0.624} &	\highlight{0.693} & \highlight{0.612} &	\highlight{0.616} &	\highlight{0.616} &	\highlight{0.686} \\
          &       & Count-Base & 0.635 & 0.639 & 0.629 & 0.703 & 0.632 & 0.630 & 0.619 & 0.697 \\
          &       & Max-Score & 0.612 & 0.595 & 0.593 & 0.677 & 0.611 & 0.598 & 0.591 & 0.673 \\
          &       & Sum-Sim & 0.635 & 0.634 & 0.627 & 0.701 & 0.629 & 0.628 & 0.618 & 0.695 \\
           \cmidrule(l){2-11}
          & \multirow{4}[0]{*}{\begin{sideways}DreamSim\end{sideways}} & C-MIR & \textbf{0.759} & \textbf{0.747} & \textbf{0.728} & \textbf{0.807} & \textbf{0.759} & \textbf{0.747} & \textbf{0.728} & \textbf{0.807} \\
          &       & \highlight{RRF} & \highlight{0.730} &	\highlight{0.705} &	\highlight{0.677} & \highlight{0.771}  & \highlight{0.738} &	\highlight{0.709} &	\highlight{0.683} &	\highlight{0.777} \\
          &       & Count-Base & 0.709 & 0.704 & 0.669 & 0.757 & 0.714 & 0.709 & 0.673 & 0.760 \\
          &       & Max-Score & 0.726 & 0.707 & 0.674 & 0.782 & 0.742 & 0.721 & 0.681 & 0.792 \\
          &       & Sum-Sim & 0.707 & 0.703 & 0.667 & 0.757 & 0.714 & 0.706 & 0.671 & 0.759 \\
          \cmidrule(l){2-11}
          & \multirow{4}[0]{*}{\begin{sideways}Swintrans.\end{sideways}} & C-MIR & \textbf{0.739} & \textbf{0.728} & \textbf{0.715} & \textbf{0.787} & \textbf{0.739} & \textbf{0.728} & \textbf{0.715} & \textbf{0.787} \\
          &       & \highlight{RRF} & \highlight{0.680} &	\highlight{0.669} &	\highlight{0.656} &	\highlight{0.736} & \highlight{0.668} &	\highlight{0.665} &	\highlight{0.650} &	\highlight{0.725} \\
          &       & Count-Base & 0.658 & 0.652 & 0.648 & 0.718 & 0.649 & 0.645 & 0.641 & 0.711 \\
          &       & Max-Score & 0.653 & 0.640 & 0.635 & 0.713 & 0.652 & 0.638 & 0.628 & 0.714 \\
          &       & Sum-Sim & 0.657 & 0.649 & 0.647 & 0.716 & 0.648 & 0.643 & 0.640 & 0.710 \\
          \hline \hline
    \multirow{12}[0]{*}{\begin{sideways}Liver\end{sideways}} & \multirow{4}[0]{*}{\begin{sideways}BioMedClip\end{sideways}} & C-MIR & \textbf{0.749} & \textbf{0.735} & \textbf{0.709} & \textbf{0.792} & \textbf{0.749} & \textbf{0.735} & 0.709 & 0.792 \\
          &       & \highlight{RRF} & \highlight{0.738} &	\highlight{0.720} &	\highlight{0.707} &  \highlight{0.789} &  \highlight{0.748} & \highlight{\textbf{0.735}} & \highlight{0.713} &	\highlight{\textbf{0.797}} \\
          &       & Count-Base & 0.742 & 0.725 & 0.707 & 0.781 & 0.740 & 0.726 & 0.717 & 0.790 \\
          &       & Max-Score & 0.723 & 0.710 & 0.694 & 0.778 & 0.731 & 0.728 & 0.707 & 0.782 \\
          &       & Sum-Sim & 0.742 & 0.724 & 0.707 & 0.782 & 0.739 & 0.725 & \textbf{0.718} & 0.790 \\ 
          \cmidrule(l){2-11}
          & \multirow{4}[0]{*}{\begin{sideways}DreamSim\end{sideways}} & C-MIR & 0.737 & 0.727 & 0.709 & 0.787 & 0.737 & 0.727 & 0.709 & 0.787 \\
          &       & \highlight{RRF} & \highlight{0.749} &	\highlight{\textbf{0.739}} & \highlight{\textbf{0.712}} &	\highlight{0.797}  &  \highlight{0.756} & \highlight{0.741} & \highlight{0.718} &	\highlight{0.802} \\
          &       & Count-Base & 0.759 & 0.736 & \textbf{0.712} & \textbf{0.807} & \textbf{0.764} & \textbf{0.742} & \textbf{0.719} & 0.810 \\
          &       & Max-Score & 0.717 & 0.704 & 0.701 & 0.768 & 0.719 & 0.708 & 0.700 & 0.771 \\
          &       & Sum-Sim & \textbf{0.760} & 0.736 & \textbf{0.712} & \textbf{0.807} & 0.763 & \textbf{0.742} & 0.717 & \textbf{0.811} \\
          \cmidrule(l){2-11}
          & \multirow{4}[0]{*}{\begin{sideways}Swintrans.\end{sideways}} & C-MIR & 0.722 & 0.715 & 0.696 & \textbf{0.784} & 0.722 & 0.715 & 0.696 & 0.784 \\
          &       & \highlight{RRF} & \highlight{\textbf{0.732}} &	\highlight{\textbf{0.722}} &	\highlight{0.698} & \highlight{0.783} &  \highlight{\textbf{0.734}} &	\highlight{\textbf{0.724}} &	\highlight{0.698} &	\highlight{\textbf{0.790}}  \\
          &       & Count-Base & 0.713 & 0.712 & 0.700 & 0.772 & 0.718 & 0.713 & \textbf{0.701} & 0.781 \\
          &       & Max-Score & 0.708 & 0.684 & 0.666 & 0.759 & 0.710 & 0.687 & 0.673 & 0.758 \\
          &       & Sum-Sim & 0.714 & 0.713 & \textbf{0.703} & 0.772 & 0.718 & 0.712 & \textbf{0.701} & 0.782 \\
          \hline \hline
    \multirow{12}[0]{*}{\begin{sideways}Lung\end{sideways}} & \multirow{4}[0]{*}{\begin{sideways}BioMedClip\end{sideways}} & C-MIR & \textbf{0.902} & \textbf{0.905} & \textbf{0.896} & \textbf{0.923} & \textbf{0.902} & \textbf{0.905} & \textbf{0.896} & \textbf{0.923} \\
          &       & \highlight{RRF} & \highlight{0.898} &	\highlight{0.893} &	\highlight{0.888} &	\highlight{0.928}  &  \highlight{0.893} & \highlight{0.889} & \highlight{0.882} & \highlight{0.919} \\
          &       & Count-Base & 0.900 & 0.887 & 0.884 & 0.921 & 0.886 & 0.880 & 0.879 & 0.912 \\
          &       & Max-Score & 0.901 & 0.888 & 0.885 & 0.921 & 0.890 & 0.885 & 0.881 & 0.911 \\
          &       & Sum-Sim & 0.900 & 0.886 & 0.884 & 0.921 & 0.886 & 0.881 & 0.879 & 0.913 \\
          \cmidrule(l){2-11}
          & \multirow{4}[0]{*}{\begin{sideways}DreamSim\end{sideways}} & C-MIR & \textbf{0.932} & \textbf{0.926} & \textbf{0.913} & \textbf{0.942} & \textbf{0.932} & \textbf{0.926} & \textbf{0.913} & \textbf{0.942} \\
          &       & \highlight{RRF} & \highlight{0.916} &	\highlight{0.902} & \highlight{0.885} &	\highlight{0.929} & \highlight{0.916} &	\highlight{0.903} & \highlight{0.885} &	\highlight{0.930} \\
          &       & Count-Base & 0.917 & 0.909 & 0.885 & 0.936 & 0.918 & 0.907 & 0.884 & 0.935 \\
          &       & Max-Score & 0.899 & 0.887 & 0.873 & 0.910 & 0.896 & 0.887 & 0.874 & 0.910 \\
          &       & Sum-Sim & 0.917 & 0.911 & 0.886 & 0.935 & 0.919 & 0.908 & 0.886 & 0.935 \\
          \cmidrule(l){2-11}
          & \multirow{4}[0]{*}{\begin{sideways}Swintrans.\end{sideways}} & C-MIR & \textbf{0.900} & \textbf{0.894} & \textbf{0.884} & \textbf{0.918} & \textbf{0.900} & \textbf{0.894} & \textbf{0.884} & \textbf{0.918} \\
          &       & \highlight{RRF} & \highlight{0.893} &	\highlight{0.887} &	\highlight{0.870} &	\highlight{0.911} & \highlight{0.890} &	\highlight{0.890} &	\highlight{0.867} &	\highlight{0.912} \\
          &       & Count-Base & 0.889 & 0.872 & 0.868 & 0.905 & 0.890 & 0.874 & 0.865 & 0.905 \\
          &       & Max-Score & 0.881 & 0.859 & 0.851 & 0.899 & 0.881 & 0.860 & 0.853 & 0.900 \\
          &       & Sum-Sim & 0.889 & 0.873 & 0.868 & 0.904 & 0.889 & 0.874 & 0.865 & 0.904 \\
          \hline \hline
    \multirow{12}[0]{*}{\begin{sideways}Pancreas\end{sideways}} & \multirow{4}[0]{*}{\begin{sideways}BioMedClip\end{sideways}} & C-MIR & \textbf{0.756} & 0.744 & \textbf{0.729} & 0.795 & \textbf{0.756} & \textbf{0.744} & 0.729 & \textbf{0.795} \\
          &       & \highlight{RRF} & \highlight{0.748} &	\highlight{0.741} &	\highlight{0.721} & \highlight{0.791} & \highlight{0.746} &	\highlight{0.739} &	\highlight{0.729} &	\highlight{0.791} \\
          &       & Count-Base & 0.753 & \textbf{0.745} & 0.724 & 0.798 & 0.745 & 0.741 & \textbf{0.731} & \textbf{0.795} \\
          &       & Max-Score & 0.738 & 0.722 & 0.708 & 0.780 & 0.723 & 0.720 & 0.712 & 0.775 \\
          &       & Sum-Sim & 0.753 & \textbf{0.745} & 0.724 & \textbf{0.799} & 0.743 & 0.739 & 0.730 & 0.794 \\
          \cmidrule(l){2-11}
          & \multirow{4}[0]{*}{\begin{sideways}DreamSim\end{sideways}} & C-MIR & 0.746 & 0.738 & 0.723 & 0.795 & 0.746 & 0.738 & 0.723 & 0.795 \\
          &       & \highlight{RRF} & \highlight{0.751} &	\highlight{0.743} &	\highlight{0.722} &	\highlight{0.795} & \highlight{0.759} &	\highlight{0.745} &	\highlight{0.726} &	\highlight{0.799} \\
          &       & Count-Base & \textbf{0.757} & \textbf{0.747} & \textbf{0.726} & \textbf{0.797} & \textbf{0.764} & 0.748 & \textbf{0.727} & \textbf{0.802} \\
          &       & Max-Score & 0.735 & 0.728 & 0.711 & 0.787 & 0.737 & 0.729 & 0.714 & 0.788 \\
          &       & Sum-Sim & 0.755 & 0.745 & 0.724 & \textbf{0.797} & 0.763 & \textbf{0.749} & 0.725 & 0.801 \\
          \cmidrule(l){2-11}
          & \multirow{4}[0]{*}{\begin{sideways}Swintrans.\end{sideways}} & C-MIR & 0.738 & 0.726 & \textbf{0.709} & 0.789 & 0.738 & 0.726 & \textbf{0.709} & 0.789 \\
          &       & \highlight{RRF} & \highlight{0.746} &	\highlight{0.728} &	\highlight{0.705} &	\highlight{0.790} & \highlight{\textbf{0.749}} &	\highlight{\textbf{0.736}} &	\highlight{\textbf{0.709}} &	\highlight{\textbf{0.794}} \\
          &       & Count-Base & \textbf{0.749} & \textbf{0.730} & 0.704 & \textbf{0.794} & 0.746 & 0.733 & \textbf{0.709} & 0.791 \\
          &       & Max-Score & 0.731 & 0.717 & 0.692 & 0.779 & 0.732 & 0.718 & 0.694 & 0.783 \\
          &       & Sum-Sim & 0.748 & 0.729 & 0.705 & 0.793 & 0.747 & 0.733 & \textbf{0.709} & 0.791 \\
    \end{tabular}%
  \label{tab:flagging combined w and wo}%
\end{table}%

\subsubsection{Organ-agnostic Database} 
\Cref{tab:unified} shows the performance of \highlight{re-ranking methods, i.e.} C-MIR \highlight{and RRF in comparison with the three vanilla aggregation methods} for four organs, three feature extractors and for the organ-agnostic database.
C-MIR \remove{improves the results}\highlight{is the best-performing method} for colon tumor flagging across all models. 
For liver tumor flagging \remove{C-MIR}\highlight{both re-ranking methods slightly} improve\remove{s} the results for BioMedClip and SwinTransformer embeddings but show\remove{s} a decline for DreamSim \highlight{embeddings}, a pattern observed similarly in organ-specific databases. 
C-MIR \remove{enhances retrieval results for DreamSim in flagging lung tumors and shows slight improvements for the SwinTransformer.}\highlight{is the best-performing method for lung tumor flagging for DreamSim embeddings but shows similar performance for other embeddings. RFF outperforms C-MIR using BioMedClip embeddings.}
For pancreas tumor flagging C-MIR improves the results for all the models \highlight{and outperforms RRF}. 

The highest AP for colon tumor flagging is $0.761$ \highlight{using C-MIR with DreamSim embeddings}\remove{for DreamSim using C-MIR}. 
\remove{For liver tumor flagging, the highest AP is $0.799$ for the count-base method using BioMedClip.}\remove{There is a huge drop in the performance of DreamSim for liver tumor flagging.}\highlight{For liver tumor flagging, RRF and C-MIR perform on par with an AP of $0.79$ using BioMedClip embeddings.}
The best-performing method for lung tumor flagging is \remove{Sum-Sim and C-MIR}\highlight{count-base and C-MIR method} using \remove{BioMedClip}\highlight{SwinTransformer embeddings} with an AP of \remove{$0.869$}\highlight{$0.88$}. 
For pancreas tumor flagging, the highest AP belongs to the \remove{DreamSim model using C-MIR}\highlight{C-MIR using DreamSim embeddings} with AP of $0.867$. 
Expanding the database allows us to observe the effect of embedding selection on individual tasks. 
\highlight{Given the correct choice of embedding for the organ-agnostic database, C-MIR shows a promising performance compared to the vanilla aggregation methods and RRF. }

\begin{table}[!htbp]
  \centering
  \aboverulesep = 0pt
    \belowrulesep = 0pt
  \tiny
  \caption{Overview of \textbf{tumor flagging and staging} results using \textbf{organ-agnostic database}. \highlight{Reported metrics represent the average values obtained from 10 experiments, each employing a different random seed for case sampling.} The bold-faced value in each sub-column shows the best method for each model.}
    \begin{tabular}{c|c|c|cccc||cccc}
          & \multirow{2}[0]{*}{Model} & \multirow{2}[0]{*}{Method} & \multicolumn{4}{c}{Flagging } & \multicolumn{4}{c}{Stagging} \\
    \multirow{13}[0]{*}{\begin{sideways}Colon\end{sideways}} &       &       & \textcolor[rgb]{ .275,  .471,  .525}{p@3} & \textcolor[rgb]{ .275,  .471,  .525}{p@5} & \textcolor[rgb]{ .275,  .471,  .525}{p@10} & AP    & \textcolor[rgb]{ .275,  .471,  .525}{p@3} & \textcolor[rgb]{ .275,  .471,  .525}{p@5} & \textcolor[rgb]{ .275,  .471,  .525}{p@10} & AP \\ \hline
          & \multirow{4}[0]{*}{\begin{sideways}BioMedClip\end{sideways}} & C-MIR & \textbf{0.621} & \textbf{0.614} & \textbf{0.606} & \textbf{0.685} & 0.524 & 0.518 & \textbf{0.514} & \textbf{0.584} \\
          &       & \highlight{RRF} & \highlight{0.596} &	\highlight{0.597} &	\highlight{0.590} &	\highlight{0.662} &	\highlight{0.517} &	\highlight{0.513} &	\highlight{0.508} &	\highlight{0.574} \\
          &       & Count-Base & 0.602 & 0.598 & 0.593 & 0.668 & \textbf{0.528} & 0.518 & 0.510 & 0.582 \\
          &       & Max-Score & 0.587 & 0.586 & 0.580 & 0.659 & 0.505 & 0.503 & 0.501 & 0.559 \\
          &       & Sum-Sim & 0.599 & 0.596 & 0.595 & 0.666 & 0.526 & \textbf{0.519} & \textbf{0.514} & 0.582 \\
          \cmidrule(l){2-11}
          & \multirow{4}[0]{*}{\begin{sideways}DreamSim\end{sideways}} & C-MIR & \textbf{0.706} & \textbf{0.696} & \textbf{0.676} & \textbf{0.761} & \textbf{0.557} & \textbf{0.556} & \textbf{0.541} & \textbf{0.642} \\
          &       & \highlight{RRF} & \highlight{0.675} &	\highlight{0.660} &	\highlight{0.643} &	\highlight{0.731} &	\highlight{0.551} &	\highlight{0.541} &	\highlight{0.530} &	\highlight{0.624}  \\
          &       & Count-Base & 0.660 & 0.653 & 0.638 & 0.718 & 0.545 & 0.539 & 0.528 & 0.617 \\
          &       & Max-Score & 0.680 & 0.659 & 0.636 & 0.736 & 0.547 & 0.533 & 0.520 & 0.622 \\
          &       & Sum-Sim & 0.662 & 0.656 & 0.639 & 0.721 & 0.547 & 0.542 & 0.530 & 0.620 \\
          \cmidrule(l){2-11}
          & \multirow{4}[0]{*}{\begin{sideways}Swintrans.\end{sideways}} & C-MIR & \textbf{0.681} & \textbf{0.670} & \textbf{0.665} & \textbf{0.738} & \textbf{0.555} & \textbf{0.545} & \textbf{0.542} & \textbf{0.619} \\
          &       & \highlight{RRF} & \highlight{0.637} &	\highlight{0.633} &	\highlight{0.628} & \highlight{0.698} &	\highlight{0.535} &	\highlight{0.532} &	\highlight{0.527} &	\highlight{0.597} \\
          &       & Count-Base & 0.614 & 0.620 & 0.624 & 0.684 & 0.526 & 0.528 & 0.526 & 0.594 \\
          &       & Max-Score & 0.625 & 0.620 & 0.617 & 0.696 & 0.525 & 0.521 & 0.515 & 0.594 \\
          &       & Sum-Sim & 0.615 & 0.621 & 0.626 & 0.684 & 0.530 & 0.532 & 0.529 & 0.594 \\

          \hline \hline
    \multirow{12}[0]{*}{\begin{sideways}Liver\end{sideways}} & \multirow{4}[0]{*}{\begin{sideways}BioMedClip\end{sideways}} & C-MIR & \textbf{0.747} & \textbf{0.739} & 0.712 & 0.798 & 0.600 & \textbf{0.600} & 0.589 & \textbf{0.674} \\
          &       & \highlight{RRF} & \highlight{0.744} &	\highlight{0.734} &	\highlight{\textbf{0.713}} &	\highlight{\textbf{0.799}} &	\highlight{0.594} &	\highlight{0.596} &	\highlight{\textbf{0.591}} &	\highlight{0.673} \\
          &       & Count-Base & 0.744 & 0.732 & 0.708 & \textbf{0.799} & 0.592 & 0.589 & 0.582 & 0.670 \\
          &       & Max-Score & 0.739 & 0.724 & 0.698 & 0.786 & \textbf{0.606} & 0.599 & 0.585 & 0.670 \\
          &       & Sum-Sim & 0.735 & 0.721 & 0.700 & 0.792 & 0.583 & 0.578 & 0.574 & 0.663 \\
          \cmidrule(l){2-11}
          & \multirow{4}[0]{*}{\begin{sideways}DreamSim\end{sideways}} & C-MIR & 0.593 & 0.569 & 0.538 & 0.671 & 0.382 & 0.369 & 0.358 & 0.487 \\
          &       & \highlight{RRF} & \highlight{0.620} &	\highlight{0.588} &	\highlight{0.540} &	\highlight{0.692} & \highlight{0.388} &	\highlight{0.378} &	\highlight{0.359} &	\highlight{0.492} \\
          &       & Count-Base & 0.632 & 0.590 & \textbf{0.543} & 0.705 & 0.397 & 0.376 & 0.359 & 0.505 \\
          &       & Max-Score & 0.587 & 0.550 & 0.521 & 0.651 & 0.386 & 0.367 & 0.354 & 0.474 \\
          &       & Sum-Sim & \textbf{0.635} & \textbf{0.592} & 0.545 & \textbf{0.707} & \textbf{0.402} & \textbf{0.379} & \textbf{0.361} & \textbf{0.507} \\
          \cmidrule(l){2-11}
          & \multirow{4}[0]{*}{\begin{sideways}Swintrans.\end{sideways}} & C-MIR & \textbf{0.708} & \textbf{0.687} & 0.663 & \textbf{0.759} & \textbf{0.559} & 0.546 & \textbf{0.533} & \textbf{0.635} \\
          &       & \highlight{RRF} & \highlight{0.706} &	\highlight{0.686} &	\highlight{0.660} &	\highlight{0.759} &	\highlight{0.559} &	\highlight{\textbf{0.548}} &	\highlight{0.527} 	 & \highlight{0.638}  \\
          &       & Count-Base & 0.695 & \textbf{0.687} & \textbf{0.664} & 0.754 & 0.540 & 0.539 & 0.526 & 0.630 \\
          &       & Max-Score & 0.667 & 0.651 & 0.635 & 0.731 & 0.545 & 0.537 & 0.525 & 0.626 \\
          &       & Sum-Sim & 0.692 & 0.685 & 0.663 & 0.754 & 0.538 & 0.537 & 0.524 & 0.630 \\

          \hline \hline 
    \multirow{12}[0]{*}{\begin{sideways}Lung\end{sideways}} & \multirow{4}[0]{*}{\begin{sideways}BioMedClip\end{sideways}} & C-MIR & 0.829 & 0.821 & 0.810 & 0.857 & 0.431 & 0.436 & 0.444 & 0.539 \\
          &       & \highlight{RRF} & \highlight{0.825} &	\highlight{\textbf{0.829}} &	\highlight{0.814} &	\highlight{\textbf{0.872}} &	\highlight{0.439} &	\highlight{0.441} &	\highlight{0.442} &	\highlight{0.546} \\
          &       & Count-Base & 0.832 & 0.826 & \textbf{0.817} & 0.868 & 0.439 & 0.436 & \textbf{0.446} & \textbf{0.550} \\
          &       & Max-Score & 0.832 & 0.826 & 0.811 & 0.857 & \textbf{0.455} &\textbf{ 0.451} & 0.441 & 0.541 \\
          &       & Sum-Sim & \textbf{0.834} & 0.827 & \textbf{0.817} & 0.869 & 0.440 & 0.438 & \textbf{0.446} & \textbf{0.550} \\
          \cmidrule(l){2-11}
          & \multirow{4}[0]{*}{\begin{sideways}DreamSim\end{sideways}} & C-MIR & \textbf{0.821} & \textbf{0.800} & \textbf{0.780} & \textbf{0.843} & \textbf{0.513} & \textbf{0.500} & \textbf{0.491} & \textbf{0.588} \\
          &       & \highlight{RRF} & \highlight{0.808} &	\highlight{0.789} &	\highlight{0.749} &	\highlight{0.829} &	\highlight{0.510} &	\highlight{0.497} &	\highlight{0.477} &	\highlight{0.583} \\
          &       & Count-Base & 0.817 & 0.794 & 0.757 & 0.837 & 0.511 & 0.494 & 0.478 & 0.584 \\
          &       & Max-Score & 0.779 & 0.755 & 0.724 & 0.803 & 0.486 & 0.483 & 0.466 & 0.559 \\
          &       & Sum-Sim & 0.819 & 0.796 & 0.758 & 0.838 & \textbf{0.513} & 0.496 & 0.481 & 0.584 \\
          \cmidrule(l){2-11}
          & \multirow{4}[0]{*}{\begin{sideways}Swintrans.\end{sideways}} & C-MIR & 0.860 & \textbf{0.857} & \textbf{0.824} & 0.881 & 0.483 & 0.481 & \textbf{0.469} & 0.577 \\
          &       & \highlight{RRF} & \highlight{0.856} &	\highlight{0.846} &	\highlight{0.815} &	\highlight{0.879} &	\highlight{0.483} & \highlight{0.484} &	\highlight{0.469} &	\highlight{0.579} \\
          &       & Count-Base & 0.864 & 0.850 & 0.822 & \textbf{0.882} & 0.485 & 0.483 & 0.467 & \textbf{0.583} \\
          &       & Max-Score & 0.830 & 0.813 & 0.786 & 0.854 & 0.474 & \textbf{0.469} & 0.453 & 0.571 \\
          &       & Sum-Sim & \textbf{0.867} & 0.849 & 0.822 & 0.881 & \textbf{0.486} & 0.484 & 0.468 & \textbf{0.583} \\

          \hline \hline
    \multirow{12}[0]{*}{\begin{sideways}Pancreas\end{sideways}} & \multirow{4}[0]{*}{\begin{sideways}BioMedClip\end{sideways}} & C-MIR & \textbf{0.793} & \textbf{0.782} & \textbf{0.765} & \textbf{0.828} & \textbf{0.591} & \textbf{0.583} & \textbf{0.569} & \textbf{0.656} \\
          &       & \highlight{RRF} & \highlight{0.771} &	\highlight{0.769}	& \highlight{0.749} &	\highlight{0.815} 	& \highlight{0.575} &	\highlight{0.572} &	\highlight{0.552} &	\highlight{0.645} \\
          &       & Count-Base & 0.776 & 0.768 & 0.754 & 0.820 & 0.584 & 0.572 & 0.559 & 0.648 \\
          &       & Max-Score & 0.742 & 0.739 & 0.721 & 0.792 & 0.544 & 0.544 & 0.532 & 0.624 \\
          &       & Sum-Sim & 0.774 & 0.767 & 0.752 & 0.818 & 0.580 & 0.569 & 0.554 & 0.645 \\
          \cmidrule(l){2-11}
          & \multirow{4}[0]{*}{\begin{sideways}DreamSim\end{sideways}} & C-MIR & 0.838 & \textbf{0.826} & \textbf{0.815} & \textbf{0.867} & 0.528 & 0.520 & 0.512 & \textbf{0.612} \\
          &       & \highlight{RRF} & \highlight{0.825} &	\highlight{0.821} &	\highlight{0.803} &	\highlight{0.861} &	\highlight{0.529} &	\highlight{0.519} &	\highlight{0.502} & \highlight{0.609}  \\
          &       & Count-Base & 0.833 & 0.819 & 0.803 & 0.860 & 0.535 & 0.525 & 0.507 & 0.609 \\
          &       & Max-Score & 0.812 & 0.804 & 0.789 & 0.847 & 0.510 & 0.500 & 0.484 & 0.591 \\
          &       & Sum-Sim & \textbf{0.840} & \textbf{0.826} & 0.810 & 0.864 & \textbf{0.541} & \textbf{0.530} & \textbf{0.513} & \textbf{0.612} \\
          \cmidrule(l){2-11}
          & \multirow{4}[0]{*}{\begin{sideways}Swintrans.\end{sideways}} & C-MIR & \textbf{0.777} & \textbf{0.762} & \textbf{0.751} & \textbf{0.815} & \textbf{0.592} & \textbf{0.579} & \textbf{0.566} & \textbf{0.660} \\
          &       & \highlight{RRF} & \highlight{0.766} &	\highlight{0.753} &	\highlight{0.734} &	\highlight{0.806} &	\highlight{0.580} &	\highlight{0.569} &	\highlight{0.553} &	\highlight{0.653} \\
          &       & Count-Base & 0.761 & 0.749 & 0.735 & 0.805 & 0.577 & 0.569 & 0.557 & 0.651 \\
          &       & Max-Score & 0.725 & 0.714 & 0.704 & 0.778 & 0.538 & 0.531 & 0.522 & 0.626 \\
          &       & Sum-Sim & 0.769 & 0.756 & 0.740 & 0.811 & 0.584 & 0.575 & 0.561 & 0.656 \\
    \end{tabular}%
  \label{tab:unified}%
\end{table}%

\subsection{Tumor Staging}
\label{sec: tumor staging}

\remove{The following sections provide the results of the tumor staging for organ-specific and organ-agnostic databases for the colon, liver, lung, and pancreas using BioMedClip, DreamSim, and SwinTransformer models as feature extractors. 
Three aggregation methods, as outlined in Section 2.6, are employed and evaluated alongside C-MIR to obtain top-k results.}

\begin{table}[!htbp]
    \centering
  \aboverulesep = 0pt
    \belowrulesep = 0pt
  \tiny
  \caption{Overview of \textbf{tumor staging} results using \textbf{organ-specific databases} with and without segmentation. \highlight{Reported metrics represent the average values obtained from 10 experiments, each employing a different random seed for case sampling.}  The bold-faced value in each sub-column shows the best method for each model.}
    \begin{tabular}{c|c|c|cccc||cccc}
          & \multirow{2}[0]{*}{Model} & \multirow{2}[0]{*}{Method} & \multicolumn{4}{c}{With Segmentation} & \multicolumn{4}{c}{Without Segmentation} \\
    \multirow{13}[0]{*}{\begin{sideways}Colon\end{sideways}} &       &       & \textcolor[rgb]{ .275,  .471,  .525}{p@3} & \textcolor[rgb]{ .275,  .471,  .525}{p@5} & \textcolor[rgb]{ .275,  .471,  .525}{p@10} & AP    & \textcolor[rgb]{ .275,  .471,  .525}{p@3} & \textcolor[rgb]{ .275,  .471,  .525}{p@5} & \textcolor[rgb]{ .275,  .471,  .525}{p@10} & AP \\ \hline
          & \multirow{4}[0]{*}{\begin{sideways}BioMedClip\end{sideways}} & C-MIR & 0.529 & \textbf{0.529} & \textbf{0.521} & \textbf{0.607} & 0.529 & \textbf{0.529} & \textbf{0.521} & \textbf{0.607} \\
          &       & \highlight{RRF} & \highlight{0.523} &	\highlight{0.519} &	\highlight{0.517} &	\highlight{0.594} &  \highlight{0.521} &	\highlight{0.521} &	\highlight{0.517} &	\highlight{0.592} \\
          &       & Count-Base & \textbf{0.534} & 0.528 & 0.519 & 0.606 & \textbf{0.535} & 0.524 & 0.519 & 0.604 \\
          &       & Max-Score & 0.506 & 0.499 & 0.499 & 0.570 & 0.508 & 0.504 & 0.499 & 0.572 \\
          &       & Sum-Sim & \textbf{0.534} & 0.528 & 0.519 & 0.606 & 0.533 & 0.524 & 0.519 & 0.602 \\
          \cmidrule(l){2-11}
          & \multirow{4}[0]{*}{\begin{sideways}DreamSim\end{sideways}} & C-MIR & \textbf{0.571} & \textbf{0.568} & \textbf{0.556} & \textbf{0.665} & 0.571 & \textbf{0.568} & \textbf{0.556} & \textbf{0.665} \\
          &       & \highlight{RRF} & \highlight{0.579} &	\highlight{0.567} &	\highlight{0.541} &	\highlight{0.658} &  \highlight{0.575} &	\highlight{0.565} &	\highlight{0.544} &	\highlight{0.659} \\
          &       & Count-Base & 0.566 & 0.561 & 0.539 & 0.644 & 0.569 & 0.565 & 0.543 & 0.642 \\
          &       & Max-Score & \textbf{0.571} & 0.554 & 0.536 & 0.653 & \textbf{0.578} & 0.565 & 0.543 & 0.661 \\
          &       & Sum-Sim & 0.565 & 0.561 & 0.539 & 0.645 & 0.569 & 0.563 & 0.542 & 0.642 \\
          \cmidrule(l){2-11}
          & \multirow{4}[0]{*}{\begin{sideways}Swintrans.\end{sideways}} & C-MIR & \textbf{0.568} & \textbf{0.558} & \textbf{0.549} & \textbf{0.648} & \textbf{0.568} & \textbf{0.558} & \textbf{0.549} & \textbf{0.648} \\
           &       & \highlight{RRF} & \highlight{0.551} &	\highlight{0.543} &	\highlight{0.535} &	\highlight{0.62} &  \highlight{0.547} &	\highlight{0.543} &	\highlight{0.534} &	\highlight{0.614} \\
          &       & Count-Base & 0.550 & 0.543 & 0.534 & 0.612 & 0.542 & 0.538 & 0.532 & 0.605 \\
          &       & Max-Score & 0.526 & 0.514 & 0.507 & 0.593 & 0.537 & 0.523 & 0.509 & 0.605 \\
          &       & Sum-Sim & 0.549 & 0.543 & 0.535 & 0.611 & 0.542 & 0.538 & 0.531 & 0.604 \\
          \hline \hline
    \multirow{12}[0]{*}{\begin{sideways}Liver\end{sideways}} & \multirow{4}[0]{*}{\begin{sideways}BioMedClip\end{sideways}} & C-MIR & \textbf{0.608} & 0.596 & \textbf{0.589} & 0.668 & \textbf{0.608} & 0.596 & 0.589 & \textbf{0.668} \\
          &       & \highlight{RRF} & \highlight{0.599} &	\highlight{0.585} &	\highlight{0.587} &	\highlight{0.669} &  \highlight{0.602} &	\highlight{0.592} &	\highlight{0.586} &	\highlight{0.664} \\
          &       & Count-Base & 0.594 & 0.583 & 0.580 & 0.662 & 0.585 & 0.577 & 0.582 & 0.658 \\
          &       & Max-Score & 0.597 & \textbf{0.598} & 0.591 & \textbf{0.670} & 0.605 & \textbf{0.601} & \textbf{0.594} & 0.665 \\
          &       & Sum-Sim & 0.592 & 0.582 & 0.581 & 0.662 & 0.584 & 0.578 & 0.582 & 0.657 \\
          \cmidrule(l){2-11}
          & \multirow{4}[0]{*}{\begin{sideways}DreamSim\end{sideways}} & C-MIR & 0.615 & 0.608 & \textbf{0.601} & 0.677 & 0.615 & 0.608 & 0.601 & 0.677 \\
          &       & \highlight{RRF} & \highlight{0.619} &	\highlight{0.613} &	\highlight{0.601} &	\highlight{0.683} &  \highlight{0.619} &	\highlight{0.616} &	\highlight{0.606} & \highlight{0.686} \\
          &       & Count-Base & \textbf{0.624} & \textbf{0.610} & 0.599 & 0.688 & \textbf{0.627} & \textbf{0.615} & \textbf{0.607} & \textbf{0.691} \\
          &       & Max-Score & 0.605 & 0.600 & 0.600 & 0.669 & 0.608 & 0.602 & 0.601 & 0.669 \\
          &       & Sum-Sim & \textbf{0.624} & 0.609 & 0.599 & \textbf{0.689} & 0.626 & \textbf{0.615} & 0.606 & \textbf{0.691} \\
          \cmidrule(l){2-11}
          & \multirow{4}[0]{*}{\begin{sideways}Swintrans.\end{sideways}} & C-MIR & 0.599 & \textbf{0.599} & \textbf{0.585} & \textbf{0.679} & 0.599 & 0.599 & \textbf{0.585} & \textbf{0.679} \\
          &       & \highlight{RRF} &  \highlight{0.608} &	\highlight{0.601} &	\highlight{0.581} &	\highlight{0.683} &  \highlight{0.612} &	\highlight{0.603} &	\highlight{0.583} &	\highlight{0.692} \\
          &       & Count-Base & 0.589 & 0.589 & 0.582 & 0.669 & 0.597 & 0.591 & 0.579 & 0.677 \\
          &       & Max-Score & \textbf{0.605} & 0.589 & 0.578 & 0.677 & \textbf{0.611} & \textbf{0.593} & 0.584 & 0.671 \\
          &       & Sum-Sim & 0.590 & 0.589 & 0.583 & 0.669 & 0.596 & 0.592 & 0.579 & 0.678 \\
          \hline \hline
    \multirow{12}[0]{*}{\begin{sideways}Lung\end{sideways}} & \multirow{4}[0]{*}{\begin{sideways}BioMedClip\end{sideways}} & C-MIR & 0.662 & 0.660 & \textbf{0.666} & 0.739 & 0.662 & 0.660 & \textbf{0.666} & \textbf{0.739} \\
          &       & \highlight{RRF} & \highlight{0.682} &	\highlight{0.659} &  \highlight{0.66} &	\highlight{0.738} &  \highlight{0.68} &	\highlight{0.657} &	\highlight{0.655} &	\highlight{0.731} \\
          &       & Count-Base & 0.676 & 0.648 & 0.653 & 0.725 & 0.663 & 0.645 & 0.651 & 0.719 \\
          &       & Max-Score & \textbf{0.685} & \textbf{0.666} & 0.664 & \textbf{0.741} & \textbf{0.676} & \textbf{0.666} & 0.660 & 0.730 \\
          &       & Sum-Sim & 0.676 & 0.647 & 0.654 & 0.725 & 0.663 & 0.646 & 0.651 & 0.720 \\
          \cmidrule(l){2-11}
          & \multirow{4}[0]{*}{\begin{sideways}DreamSim\end{sideways}} & C-MIR & \textbf{0.673} & \textbf{0.679} & \textbf{0.670} & 0.731 & 0.673 & \textbf{0.679} & \textbf{0.670} & 0.731 \\
          &       & \highlight{RRF} & \highlight{0.669} &	\highlight{0.662} & \highlight{0.657} & \highlight{0.736}  &  \highlight{0.667} & \highlight{0.666} & \highlight{0.656} &	\highlight{0.738} \\
          &       & Count-Base & 0.666 & 0.661 & 0.653 & 0.735 & \textbf{0.670} & 0.661 & 0.654 & 0.737 \\
          &       & Max-Score & 0.639 & 0.649 & 0.654 & 0.705 & 0.637 & 0.649 & 0.656 & 0.705 \\
          &       & Sum-Sim & 0.667 & 0.663 & 0.654 & \textbf{0.737} & \textbf{0.670} & 0.661 & 0.656 & \textbf{0.738} \\
          \cmidrule(l){2-11}
          & \multirow{4}[0]{*}{\begin{sideways}Swintrans.\end{sideways}} & C-MIR & \textbf{0.669} & \textbf{0.669} & \textbf{0.671} & 0.727 & \textbf{0.669} & \textbf{0.669} & \textbf{0.671} & 0.727 \\
          &       & \highlight{RRF} & \highlight{0.663} &	\highlight{0.666} &	\highlight{0.663} &	\highlight{0.72}  &  \highlight{0.663} &	\highlight{0.676} &	\highlight{0.662} &	\highlight{0.724} \\
          &       & Count-Base & 0.663 & 0.664 & 0.659 & 0.718 & 0.661 & \textbf{0.669} & 0.659 & 0.722 \\
          &       & Max-Score & 0.662 & 0.654 & 0.656 & \textbf{0.732} & 0.661 & 0.656 & 0.657 & \textbf{0.731} \\
          &       & Sum-Sim & 0.664 & 0.664 & 0.658 & 0.718 & 0.661 & \textbf{0.669} & 0.659 & 0.722 \\
          \hline \hline
    \multirow{12}[0]{*}{\begin{sideways}Pancreas\end{sideways}} & \multirow{4}[0]{*}{\begin{sideways}BioMedClip\end{sideways}} & C-MIR & \textbf{0.562} & \textbf{0.554} & \textbf{0.539} & 0.628 & \textbf{0.562} & \textbf{0.554} & 0.539 & \textbf{0.628} \\
          &       & \highlight{RRF} & \highlight{0.558} &	\highlight{0.549} & \highlight{0.534} &	\highlight{0.63} &  \highlight{0.552} &	\highlight{0.547} &	\highlight{0.54} &	\highlight{0.625}  \\
          &       & Count-Base & \textbf{0.562} & \textbf{0.554} & \textbf{0.539} & \textbf{0.633} & 0.559 & 0.549 & \textbf{0.542} & 0.627 \\
          &       & Max-Score & 0.557 & 0.541 & 0.526 & 0.626 & 0.537 & 0.537 & 0.532 & 0.617 \\
          &       & Sum-Sim & 0.560 & 0.553 & 0.537 & \textbf{0.633} & 0.556 & 0.545 & 0.539 & 0.625 \\
          \cmidrule(l){2-11}
          & \multirow{4}[0]{*}{\begin{sideways}DreamSim\end{sideways}} & C-MIR & 0.551 & 0.539 & 0.534 & 0.629 & 0.551 & 0.539 & 0.534 & 0.629 \\
          &       & \highlight{RRF} & \highlight{0.565} &	\highlight{0.552} &	\highlight{0.534} &	\highlight{0.642} &  \highlight{0.575} &	\highlight{0.554} &	\highlight{0.535} &	\highlight{0.644} \\
          &       & Count-Base & \textbf{0.572} & \textbf{0.559} & \textbf{0.540} & \textbf{0.642} & \textbf{0.575} & \textbf{0.562} & \textbf{0.539} & \textbf{0.645} \\
          &       & Max-Score & 0.544 & 0.537 & 0.521 & 0.633 & 0.547 & 0.536 & 0.520 & 0.631 \\
          &       & Sum-Sim & 0.570 & 0.557 & 0.537 & 0.641 & 0.573 & 0.562 & 0.536 & \textbf{0.645} \\
          \cmidrule(l){2-11}
          & \multirow{4}[0]{*}{\begin{sideways}Swintrans.\end{sideways}} & C-MIR & 0.557 & 0.547 & 0.528 & 0.637 & 0.557 & 0.547 & 0.528 & 0.637 \\
          &       & \highlight{RRF} & \highlight{0.564} &	\highlight{0.553} &	\highlight{0.534} &	\highlight{0.641} &  \highlight{\textbf{0.562}} &	\highlight{\textbf{0.555}} &	\highlight{0.533} &	\highlight{\textbf{0.641}} \\
          &       & Count-Base & \textbf{0.572} & \textbf{0.562} & \textbf{0.536} & \textbf{0.645} & 0.563 & \textbf{0.555} & \textbf{0.535} & 0.640 \\
          &       & Max-Score & 0.554 & 0.542 & 0.514 & 0.630 & 0.546 & 0.536 & 0.515 & 0.630 \\
          &       & Sum-Sim & 0.570 & 0.560 & 0.535 & 0.644 & 0.562 & 0.554 & 0.534 & 0.639 \\
    \end{tabular}%
  \label{tab:staging w and wo}%
\end{table}%

\subsubsection{Organ-specific Database}

\Cref{tab:staging w and wo} presents the\remove{C-MIR} performance \highlight{of re-ranking methods, i.e., C-MIR and RRF for} \remove{in} tumor staging, \remove{compared to}\highlight{in comparison with the three} \remove{various}aggregation techniques across four organs and three feature extractors for organ-specific databases, \remove{both}with and without segmentation.
C-MIR \remove{improves the results for}\highlight{has the highest performance for} colon tumor staging for all the models. 
For the staging of liver tumors, \remove{C-MIR}\highlight{re-ranking} enhances the results of BioMedClip and SwinTransformer embeddings to some extent, yet no clear, consistent trend emerges. On the other hand, DreamSim \highlight{embeddings} demonstrate a decline in performance.
For lung tumor staging, C-MIR enhances the results, specifically for BioMedClip and SwinTransformer \highlight{embeddings}, but demonstrates a decrease in performance \remove{for}\highlight{of} DreamSim \highlight{embeddings}. \highlight{RRF follows a similar trend as C-MIR.}
In the context of pancreas tumor staging, C-MIR improved the performance of BioMedClip \highlight{embeddings}, although it led to declines for DreamSim and SwinTransformer \highlight{embeddings}.
It is noteworthy that the C-MIR results are consistent for both databases, demonstrating its capability to localize relevant regions without requiring prior segmentation to choose organ slices.

C-MIR, employing DreamSim embeddings, achieved the highest AP of $0.665$ for colon tumor staging. For liver tumor staging, the highest AP is $0.689$ for the Sum-Sim method using DreamSim \highlight{embeddings} for the database with segmentation and $0.691$ for the database without segmentation. The best-performing method for lung tumor staging is C-MIR using BioMedClip \highlight{embeddings} with an AP of $0.739$ for the database without segmentation and $0.741$ using Max-Score for the database with segmentations. 
For staging pancreatic tumors, the \highlight{highest} AP is achieved by the DreamSim \remove{model}\highlight{embeddings}, using the count-based method with an AP of $0.645$ without segmentation, and by the SwinTransformer \remove{model}\highlight{embeddings}, also employing the count-based method, with an AP of $0.645$ with segmentation.

\subsubsection{Organ-agnostic Database}

\Cref{tab:unified} shows the performance \highlight{of re-ranking methods, i.e.,} C-MIR \highlight{and RRF} for tumor staging in comparison with \remove{different}\highlight{the three vanilla} aggregation methods for four organs and three feature extractors for the organ-agnostic database.
C-MIR \remove{improves the results}\highlight{is the best-performing method} for colon tumor staging for all the models. 
For liver tumor \highlight{staging}\remove{flagging C-MIR}\highlight{ re-ranking methods} improve\remove{s} the results for BioMedClip and SwinTransformer embeddings but show\remove{s} a decline for DreamSim \highlight{embeddings}, mirroring the trend observed in tumor flagging.
For staging lung tumors, C-MIR \remove{enhances}\highlight{shows the best}\remove{the} outcomes for DreamSim \highlight{embeddings} but \highlight{both re-ranking methods show declined performance} \remove{reduces the metrics} for \remove{both} BioMedClip and SwinTransformer \highlight{embeddings}.
In pancreas tumor staging, C-MIR \highlight{is the best performing method for} \remove{enhances}BioMedClip\remove{'s} and \highlight{SwinTransformer embeddings}\remove{results}, but \remove{reduces}the performance \highlight{drops} for DreamSim and SwinTransformer \highlight{embeddings}. 

C-MIR achieved the best AP of $0.642$ for colon tumor staging using DreamSim embeddings. 
For liver tumor staging, the C-MIR method using BioMedClip embeddings achieves the highest AP of $0.674$. 
The best-performing method for lung tumor staging is C-MIR using DreamSim embeddings, with an AP of $0.588$.
In the staging of pancreatic tumors, C-MIR utilizing \remove{BioMedClip}\highlight{SwinTransformer }embeddings achieves the highest AP of \remove{$0.656$}\highlight{$0.660$}. 
\remove{As we transition to a larger database, the C-MIR method demonstrates an enhancement in average precision\remove{s}.}
\highlight{In summary, C-MIR achieves the best performance for tumor staging across all four anatomical sites.}

\subsection{Statistical Analysis}
\label{sec: stastical analysis}

\remove{Section 4} \Cref{sec: tumor flagging}, and \Cref{sec: tumor staging} show\highlight{ed} that the C-MIR method exhibits varying performance levels when applied to different organs and datasets. 
Although C-MIR enhances tumor flagging and staging for specific organs and models, there are cases, especially with DreamSim \highlight{embeddings}, where the performance drops.
These variations highlight the need for statistical analysis to evaluate the significance of the findings.
To this end, we employed a two-sided Wilcoxon signed-rank test to assess the average precision of the C-MIR method against the best method for each database.
The statistical test serves two purposes: First, it evaluates whether instances where C-MIR outperforms other methods reflect statistically significant improvements rather than random chance. Second, it assesses whether any observed declines in C-MIR's performance, indicated by a lower average compared to other methods, are statistically significant. This approach aims to ensure that any changes in performance metrics are meaningful and reliable, rather than random \remove{fluctuations}\highlight{variations}.

\subsubsection{Tumor Flagging}

\Cref{tab:ap flagging pvalues}\highlight{ contains}\remove{demonstrated} the respective p-values for tumor flagging. 
The C-MIR method shows statistically significant improvements \highlight{over the three vanilla aggregation methods and the RRF re-ranking }in colon flagging across all databases and models, highlighting its robustness in this application.
For liver tumor flagging using BioMedClip and SwinTransformer embeddings, \remove{C-MIR}\highlight{re-ranking methods do not show statistically significant improvements despite improvements in average APs.} \remove{shows statistically significant improvements for the organ-specific database with segmentation for BioMedClip and SwinTansformer embeddings. Yet, despite improvements in APs as seen in Table 4 and Table 5, the effect of C-MIR on the organ-specific database without segmentation and the organ-agnostic database is not statistically significant.}For the DreamSim model, \remove{C-MIR}\highlight{re-ranking} even declines the performance. 
C-MIR demonstrates statistically significant enhancements in lung tumor flagging for DreamSim \remove{and SwinTransformer models}\highlight{embeddings} when applied to organ-specific databases. 
However, the performance of C-MIR for the organ-agnostic database in combination with the BioMedClip \remove{model}\highlight{embeddings} shows a decline. 
C-MIR shows a subtle improvement in flagging pancreas tumors, only enhancing the results of BioMedClip \remove{models}\highlight{embeddings} in the organ-agnostic database, while its performance decreases for DreamSim \highlight{embeddings} in the organ-specific database without segmentation. The other differences in APs are not statistically significant.

\begin{table} [!htbp]
  \centering
  \caption{Wilcoxon test on average precision for \textbf{tumor flagging} of C-MIR versus the best-performing method. The bold-faced values highlight the p-values smaller than 0.05. The underlined methods indicate where C-MIR, on average, performed worse than the specified method. In all other instances, C-MIR demonstrated improvements in average AP scores in \remove{Section 4} \Cref{sec: tumor flagging} and \Cref{sec: tumor staging}.  }
  \tiny
    \addtolength{\tabcolsep}{-.2em}
    \begin{tabular}{c|c|c|c|c||c|c}
    \multirow{4}[0]{*}{Organ} & \multirow{4}[0]{*}{Model} & \multirow{4}[0]{*}{Method} & Organ-specific  & Organ-specific  & \multirow{4}[0]{*}{Method} & Organ-agnostic  \\
          &       &       & Database w.  &  Database wo.  &        &  Database \\ 
           &       &       & Segmentation & Segmentation &       & (P-value) \\ 
           &       &       & (P-value) & (P-value) &       & \\\hline
    \multirow{3}[0]{*}{colon} & BioMedClip & Count-base & \textbf{.002} & \textbf{.002} & Count-base & \textbf{.004}  \\ 
          & DreamSim & \remove{Count-base}\highlight{RRF} & \textbf{\remove{.002}} \textbf{\highlight{.003}} & \textbf{\remove{.002}}  \textbf{\highlight{.048}} &\remove{Sum-Sim}\highlight{RRF} & \textbf{\remove{.002}} \textbf{\highlight{.001}} \\
          & SwinTrans. & \remove{Count-base}\highlight{RRF} & \textbf{\remove{.002}} \textbf{\highlight{.001}} & \textbf{\remove{.002}}  \textbf{\highlight{.001}}  & \remove{Count-base}\highlight{RRF}  & \remove{\textbf{.002}} \textbf{\highlight{.001}}  \\ \hline
    \multirow{3}[0]{*}{liver} & BioMedClip & \remove{Count-base}\highlight{RRF} & \textbf{\remove{.014}}\highlight{.275} & \remove{.846}\highlight{.322}  & \remove{Count-base}\highlight{RRF} & \remove{1.000}\highlight{.769} \\
          & DreamSim & \underline{Sum-Sim} & \textbf{.002} & \textbf{.002} & \underline{Sum-Sim} & \textbf{.002} \\
          & SwinTrans. & \remove{Count-base}\highlight{RRF} & \textbf{\remove{.020}}\highlight{.921} & \remove{.625}\highlight{.160} & Count-base & .275 \\ \hline
    \multirow{3}[0]{*}{lung} & BioMedClip & \remove{Count-base}\highlight{RRF} & \remove{.695}\highlight{.431} & \remove{.275}\highlight{.695} & \underline{\remove{Sum-sim}\highlight{RRF}} & \textbf{\remove{.027}\highlight{.009}} \\ 
          & DreamSim & Count-base & \textbf{.037} & \textbf{.049} & Sum-Sim & .232 \\
          & SwinTrans. & \remove{Count-base}\highlight{RRF} & \textbf{\remove{.004}}\highlight{.083} & \textbf{\remove{.004}}\highlight{1.000} & Count-base & .922 \\ \hline
    \multirow{3}[0]{*}{pancreas} & BioMedClip & Count-base & .232 & .770 & \remove{Count-base}\highlight{RRF} & \textbf{\remove{.002}\highlight{.001}} \\ 
          & DreamSim & \underline{Count-base} & .160 & \textbf{.014} & Sum-Sim & .557 \\
          & SwinTrans. & \underline{Count-base} & .492 & .695 & Sum-Sim & .557 \\ \hline
    \end{tabular}%
  \label{tab:ap flagging pvalues}%
\end{table}%

\begin{table}[!htbp]
  \centering
  \caption{Wilcoxon test on average precision for \textbf{tumor staging} of C-MIR versus the best-performing method. The bold-faced values highlight the p-values smaller than 0.05. The underlined methods indicate where C-MIR, on average, performed worse than the specified method. In all other instances, C-MIR demonstrated improvements in average AP scores in \remove{Section 4} \Cref{sec: tumor flagging} and \Cref{sec: tumor staging}.}
  
     \tiny
    \addtolength{\tabcolsep}{-.2em}
    \begin{tabular}{c|c|c|c|c||c|c}
    \multirow{4}[0]{*}{Organ} & \multirow{4}[0]{*}{Model} & \multirow{4}[0]{*}{Method} & Organ-specific  & Organ-specific  & \multirow{4}[0]{*}{Method} & Organ-agnostic  \\
          &       &       & Database w.  &  Database wo.  &        &  Database \\ 
           &       &       & Segmentation & Segmentation &       & (P-value) \\ 
           &       &       & (P-value) & (P-value) &       & \\\hline
    \multirow{3}[0]{*}{colon} & BioMedClip & Count-base & .846 & .432 & Count-base & 1.000 \\ 
          & DreamSim & \remove{Count-base}\highlight{RRF} & \textbf{\remove{.020}}\highlight{.193} & \textbf{\remove{.027}}\highlight{.232} & \remove{Sum-Sim}\highlight{RRF} & \textbf{\remove{.002}\highlight{.001}} \\
          & SwinTrans. & \remove{Count-base}\highlight{RRF} & \textbf{\remove{.004}\highlight{.001}} & \textbf{\remove{.002}\highlight{.001}} & \remove{Count-base}\highlight{RRF} & \textbf{\remove{.004}\highlight{.005}} \\ \hline
    \multirow{3}[0]{*}{liver} & BioMedClip & Count-base & .322 & .131 & \remove{Count-base}\highlight{RRF} & \textbf{\remove{.038}}\highlight{1.000} \\
          & DreamSim & \underline{Count-base} & .105 & \textbf{.027} & \underline{Sum-Sim} & \textbf{.020} \\
          & SwinTrans. & \remove{Count-base} \underline{\highlight{RRF}} & \remove{.275}\highlight{.625}  & \remove{.846}\highlight{.160} & \remove{Count-base} \underline{\highlight{RRF}} & \remove{.275}\highlight{.625}\\ \hline
    \multirow{3}[0]{*}{lung} & BioMedClip & Count-base & .275 & .131 & \underline{Sum-Sim} & \textbf{.049} \\
          & DreamSim & \underline{Sum-Sim} & .432 & .193 & Sum-Sim & .770 \\
          & SwinTrans. & Count-base & .375 & .846 & \underline{Count-base} & .625 \\ \hline
    \multirow{3}[0]{*}{pancreas} & BioMedClip & Count-base & .232 & .922 & Count-base & \textbf{.002} \\
          & DreamSim & Count-base & \textbf{.006} & \textbf{.004} & \underline{Sum-Sim} & .846 \\
          & SwinTrans. & \underline{Count-base} & .105 & .557 & Sum-Sim & .557 \\ \hline
    \end{tabular}%
  \label{tab:ap staging pvalues}%
\end{table}%


\subsubsection{Tumor Staging}

\Cref{tab:ap staging pvalues} presents p-values from the two-sided Wilcoxon signed-rank test comparing the average precision of the C-MIR method with the top-performing method for each database in tumor staging.
In colon tumor staging, the C-MIR method demonstrates statistically significant enhancements using the DreamSim \highlight{embeddings in organ-agnostic database} and SwinTransformer embeddings across all databases. 
In liver tumor staging, a similar trend as flagging is noted: C-MIR reduces performance with DreamSim embeddings\remove{ but enhances it with BioMedClip embeddings using the organ-specific database}. For other models, although there was an increase in AP, these improvements are not statistically significant.
Lung tumor staging is particularly difficult with no improvement in overall performance using C-MIR \highlight{or RRF}. 
The C-MIR method shows statistically significant improvements for pancreatic tumor staging, particularly with the DreamSim embedding and organ-specific database. 
For the organ-agnostic database, C-MIR shows statistically significant improvements only for the BioMedClip embeddings. 

\section{Discussion}

In this study, we conducted a comprehensive evaluation of CBIR systems for 3D medical image retrieval, with a particular emphasis on tumor flagging and staging. 
Our work builds upon existing methods, extending the evaluation to databases of varying configurations. 
We introduced the \highlight{novel} ColBERT-Inspired Medical \highlight{Image} Retrieval and Re-Ranking (C-MIR) method, which takes into account the information of the whole volume for re-ranking the top-k retrieved cases.
\highlight{We compared C-MIR with a meta re-ranking method and three vanilla retrieval methods that do not re-rank.}

\subsection{Performance of C-MIR}
\remove{Our findings have shown that the C-MIR maintains steady performance across databases with varying content. Specifically, the performance was steady, independent of whether an image-only and a segmentation-mask-enhanced database had been subjected. The latter was intended to aid the precise selection of slices from relevant organs. This consistency suggests that the C-MIR method can effectively localize the relevant regions within the images, thereby potentially serving as a viable alternative to running AI-based segmentation for all volumes in a database. This has high relevance because running AI-based segmentation occurs significant compute costs (which should ideally be avoided) and in theory need to be repeated with any update of a new segmentation model. We showed that a prior segmentation to filter slices for creating organ-specific databases does not add value in the context of our CBIR system. This finding is particularly important as it aligns with real-world clinical practices where computational resources are limited, and segmentation of large datasets is often not feasible.}

\highlight{Our findings demonstrate that C-MIR maintains consistent performance across databases, regardless of whether the images are retrieved from an image-only or a segmentation-mask-enhanced database (the latter being designed for precise organ-specific slice selection). This indicates that the additional 3D context information encoded in C-MIR's similarity matrices improves localization of relevant anatomical regions without requiring prior localization, e.g., by segmentation. Since C-MIR only relies on slice embeddings that are needed for the vector similarity search anyway, this method is a computationally efficient alternative to search systems that rely on prior image segmentation or related types of computationally expensive data enrichment. This advantage is particularly evident when dealing with large volumetric image databases. 
In contrast, retrieval methods lacking 3D image context show performance variability (See Appendix} \ref{secA2}\highlight{ for a detailed statistical analysis). For these approaches, using an organ-specific database with pre-selected slices that exclude non-informative background, e.g., by organ segmentation (similar to the organ-specific database with segmentation), can improve tumor flagging, depending on the embedding or aggregation method. C-MIR provides a mean to eliminate this dependency, achieving equivalent performance while avoiding any kind of slices pre-filtering, e.g., by utilizing a segmentation model. This is a significant advantage in resource-constrained clinical settings where large-scale segmentation is often impractical.}

We showed that the C-MIR method can \highlight{be used effectively in the context of CBIR for medical image data in the presence of pathologies (here, tumors). Specifically, C-MIR could} improve tumor flagging,\remove{especially} in colon and lung cases. \highlight{Given the correct choice of embedding C-MIR performs well for liver and pancreas tumor flagging as well (best or second-best).}
\remove{However,}\highlight{It is noteworthy that} the effectiveness of this method \highlight{as well as other methods} varied depending on the embedding model, especially for larger databases\remove{, indicating the importance of embedding choice in achieving optimal retrieval results.}. Conceptually, this is not a weakness of the C-MIR framework itself, as the embedding generation can easily be updated \highlight{at any time} to the \remove{current}\highlight{latest available} state-of-the-art \highlight{models}. In other words, as increasingly more \remove{accurate pre-trained models (}foundation models\remove{)}\highlight{ with the capability to generalize on broader tasks} become available in the future, \remove{also} medical image retrieval will \highlight{also} become more accurate.
For tumor staging, the results were more variable, suggesting that further refinement of these methods is necessary to improve performance. C-MIR had the highest APs for tumor staging for all the organs in comparison with \remove{simple}\highlight{vanilla} aggregation methods in the organ-agnostic database. Nevertheless, the results revealed areas where the method did not achieve any significant improvements, indicating a need for further research.

\subsection{Challenges in Tumor Staging} 
\remove{Tumor staging presents a unique challenge due to its reliance on precise image dimensions, spacing, and other original characteristics that may be altered during preprocessing, particularly when resizing images for feature extraction.}
\highlight{Automated tumor staging faces significant challenges due to the clinical staging requirements and workflows. Tumor staging relies on precise, scale-dependent features such as absolute physical size (e.g., tumor diameter in millimeters) and anatomical context, which clinicians derive from raw medical images and image metadata like pixel spacing and slice thickness.}
This aspect of CBIR systems warrants additional research to ensure that critical image details are preserved and accurately represented in the retrieval process. In contrast, tumor flagging generally yields better results since it primarily focuses on the presence of tumors rather than their size and other detailed characteristics. When moving from flagging to staging, the importance of these detailed characteristics becomes increasingly significant, as staging requires a more nuanced analysis that takes into account the exact dimensions and growth patterns of the tumor. Hence, while flagging can be effectively handled by the current CBIR approach, staging necessitates advancements in preserving and utilizing the full range of image details to improve retrieval accuracy. To enhance the effectiveness of tumor staging, future studies should focus on utilizing higher-resolution images and fine-grained details\highlight{, using multi-resolution approaches or leveraging anatomical landmarks (e.g., vertebrae, blood vessels) as intrinsic reference points to estimate tumor size proportionally.}

\subsection{Limitations and Potentials of the C-MIR Method} 

Furthermore, it is crucial to acknowledge the limitations of re-ranking methods. Since re-ranking only modifies the order of the top retrieved cases, its effectiveness is inherently dependent on the initial retrieval quality. If the first retrieval does not return relevant cases among the top results, the effectiveness of re-ranking solutions, including C-MIR, is limited. This highlights the importance of robust initial retrieval mechanisms to fully leverage the benefits of re-ranking methods such as C-MIR.
While we utilized C-MIR for re-ranking in this study, it is worth noting that the C-MIR approach, with its full embedding matrix, could also be applied as a primary retrieval system. However, such an application would require loading the matrices of volumes into memory, which is feasible only for small datasets due to the substantial computational resources it demands. Future research can focus on exploring the scalability of C-MIR and its application to larger datasets for image retrieval, as well as enhancing the initial retrieval mechanisms to improve overall re-ranking performance.

\subsection{Scalability and Computational Efficiency}

\highlight{C-MIR is used as a re-ranking method here to ensure scalability for large datasets. Based on an initial top-k similarity search, C-MIR is applied only to these top-k candidate volumes. It only relies on the vector embeddings related to the slices of the query and the top-k image volumes. This significantly reduces the computational burden. For example, with an embedding dimension of 1024, a query volume of 300 slices, and re-ranking the top 20 candidate volumes (each with 250-500 slices), matrix multiplications overall require approximately $6.14$B FLOPs (307.2M FLOPs per each matrix multiplication) and $<15$MB of GPU memory (assuming 32-bit floating point precision). Modern GPUs and CPUs can easily handle this workload in milliseconds, and the small memory footprint allows for efficient processing.

The re-ranking approach ensures that the computational cost scales primarily with the number of top-k candidates considered for re-ranking, not the overall database size. Furthermore, the computation for each candidate volume is independent, allowing for efficient parallelization via batch processing. This makes C-MIR a scalable solution for improving retrieval accuracy in large-scale datasets, maintaining robust performance even as the dataset grows, while remaining computationally tractable.}

\subsection{Future Directions for Re-Ranking Methods}
 \highlight{Most existing re-ranking approaches in the literature are developed for text retrieval or 2D image domains. 
 Future work could focus on adapting these methods to handle the unique challenges of volumetric data, particularly the inherent variability in slice counts across medical volumes. 
 Such adaptations would need to address computational efficiency and memory constraints inherent to 3D data. Comparative evaluation of these adapted methods against C-MIR would help identify optimal strategies for volumetric re-ranking, particularly in scenarios requiring fine-grained similarity assessment across variable-length volumes. This exploration could also reveal whether techniques successful in text/2D domains (e.g., late interaction, cross-attention mechanisms) generalize effectively to 3D medical imaging.
 
 Furthermore, a critical direction for future research is the validation of these re-ranking methods on independent, external datasets. 
 This is essential to assess their generalizability to real-world clinical data and to ensure that the observed performance gains are not specific to the public dataset used in this study. Such external validation should ideally involve datasets from multiple institutions with varying imaging protocols and patient populations to provide a robust assessment of the methods' clinical utility.}

\section{Conclusion}

In this study, we introduced a novel re-ranking and retrieval approach called C-MIR, inspired by the principles of ColBERT, \highlight{where 2D slices (analogous to words) and 3D volumes (analogous to passages) are encoded into multi-vector representations using pre-trained vision models. By computing maximum cosine similarities between query slices and all slices in retrieved volumes, C-MIR leverages the inherent three-dimensional spatial context of radiological data to refine relevance rankings}. \remove{This method effectively utilizes volumetric data, fully leveraging the inherent three-dimensional structure of radiology images to enhance retrieval relevance.} We showed that C-MIR can be used in the context of CBIR retrieval and improve the outcome, especially in tumor flagging \remove{specifically improve tumor flagging}. Additionally, our evaluation demonstrates that C-MIR can effectively localize regions of interest \highlight{by incorporating context similarity}. \remove{ C-MIR is computationally efficient and can be applied to larger, unannotated datasets.However, the challenge of accurately staging tumors using CBIR systems remains an area for further research.}\highlight{The proposed method demonstrates computational efficiency and scalability for large, unannotated datasets, offering practical value for real-world clinical applications. While the method reliably flags tumor presence in retrieved cases—a critical first step for diagnostic workflows—the tumor stage of retrieved instances showed variability across experiments. This indicates that while C-MIR effectively identifies tumor-afflicted cases, refining its ability to match precise staging criteria remains an important focus for future work.} This study establishes a basis for future research to create more robust and efficient retrieval techniques by leveraging an existing \remove{benchmark while eliminating the necessity for}\highlight{method without requiring} prior segmentation \remove{and dependence on}\highlight{or} organ-specific databases. Our findings contribute to the growing body of literature on CBIR in the medical domain, emphasizing the urgent need for reliable and efficient retrieval methods that can be seamlessly integrated into clinical workflows.

\begin{appendices}

\section{Qualitative Examples}\label{secA1}

\begin{figure}[!htpb]
    \centering
    \includegraphics[width=0.98\textwidth, trim=0 3cm 0 1cm, clip]{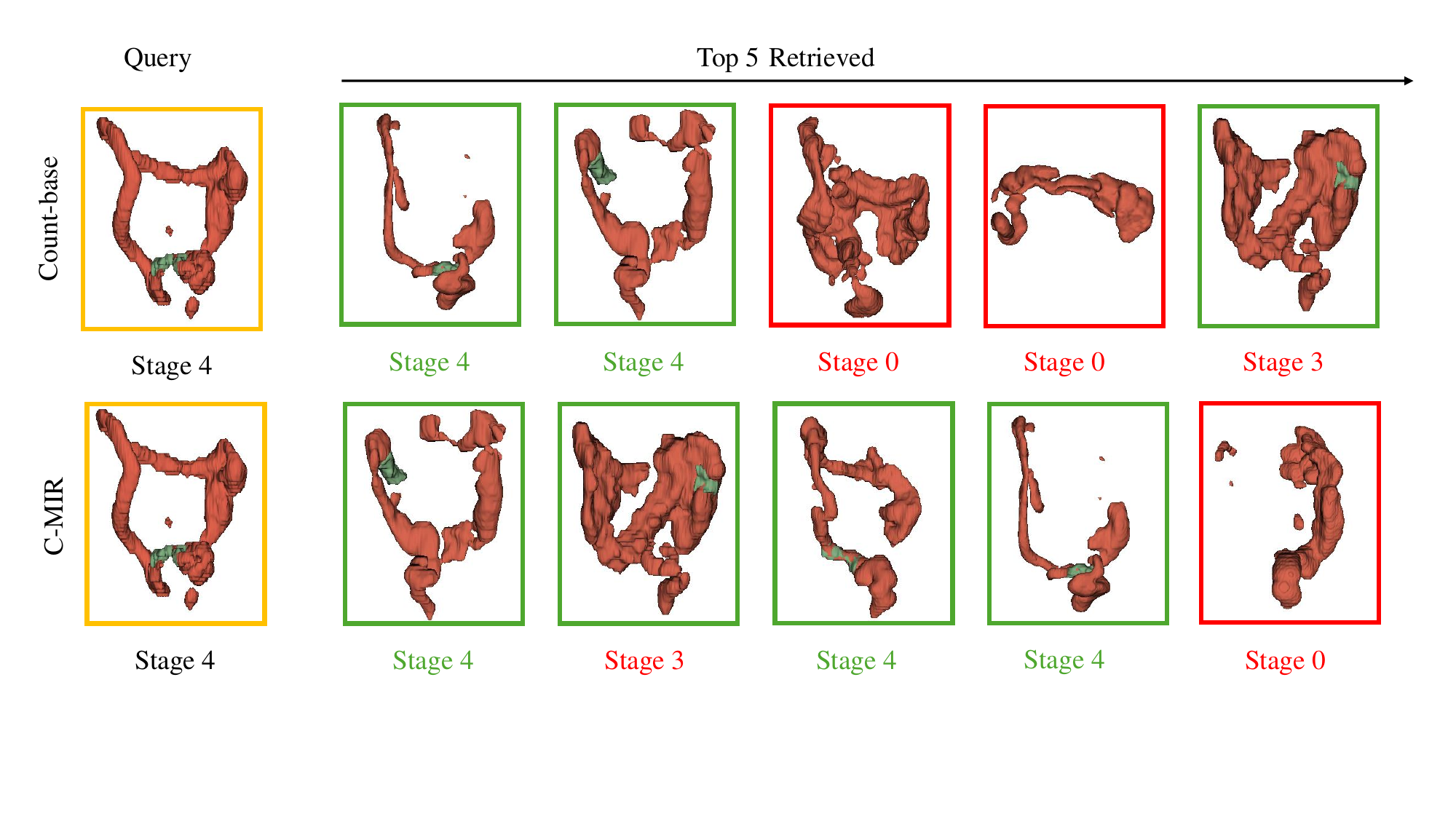}
    \caption{Visual representation of retrieval outcomes for one colon case, with the top five retrieved results. The colon segmentation mask is shown in red, and the tumors are denoted in green. Green boxes indicate accurate tumor flags, while red boxes indicate failures. Stage numbers below each instance show the actual and matched stages, with green for correct and red for incorrect matches. C-MIR improved tumor flagging and tumor staging for the top five retrieved cases.}
    \label{fig:quantitative colon}
\end{figure}

\begin{figure}[!htpb]
    \centering
    \includegraphics[width=0.98\textwidth, trim=0 3cm 0 1cm, clip]{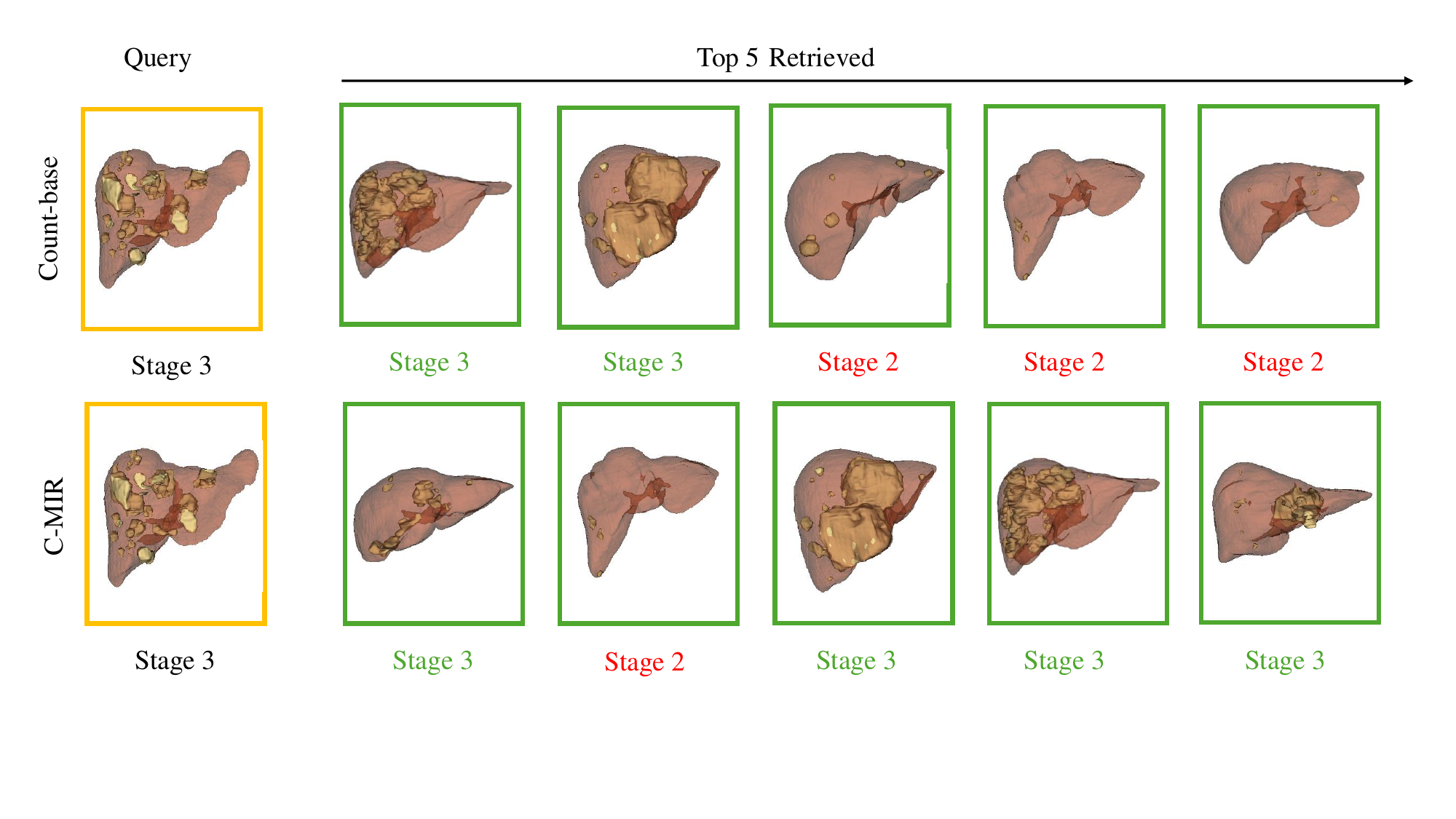}
    \caption{Visual representation of retrieval outcomes for one liver case, with the top five retrieved results. The liver segmentation mask is shown in brown, and the tumors are denoted in yellow. Green boxes indicate accurate tumor flags, while red boxes indicate failures. Stage numbers below each instance show the actual and matched stages, with green for correct and red for incorrect matches. C-MIR improved tumor staging for the top five retrieved cases. Both count-base and C-MIR methods indicate a perfect tumor flagging score for the top five retrieved cases. }
    \label{fig:quantitative liver}
\end{figure}

\begin{figure}[!htpb]
    \centering
    \includegraphics[width=0.98\textwidth, trim=0 3cm 0 1cm, clip]{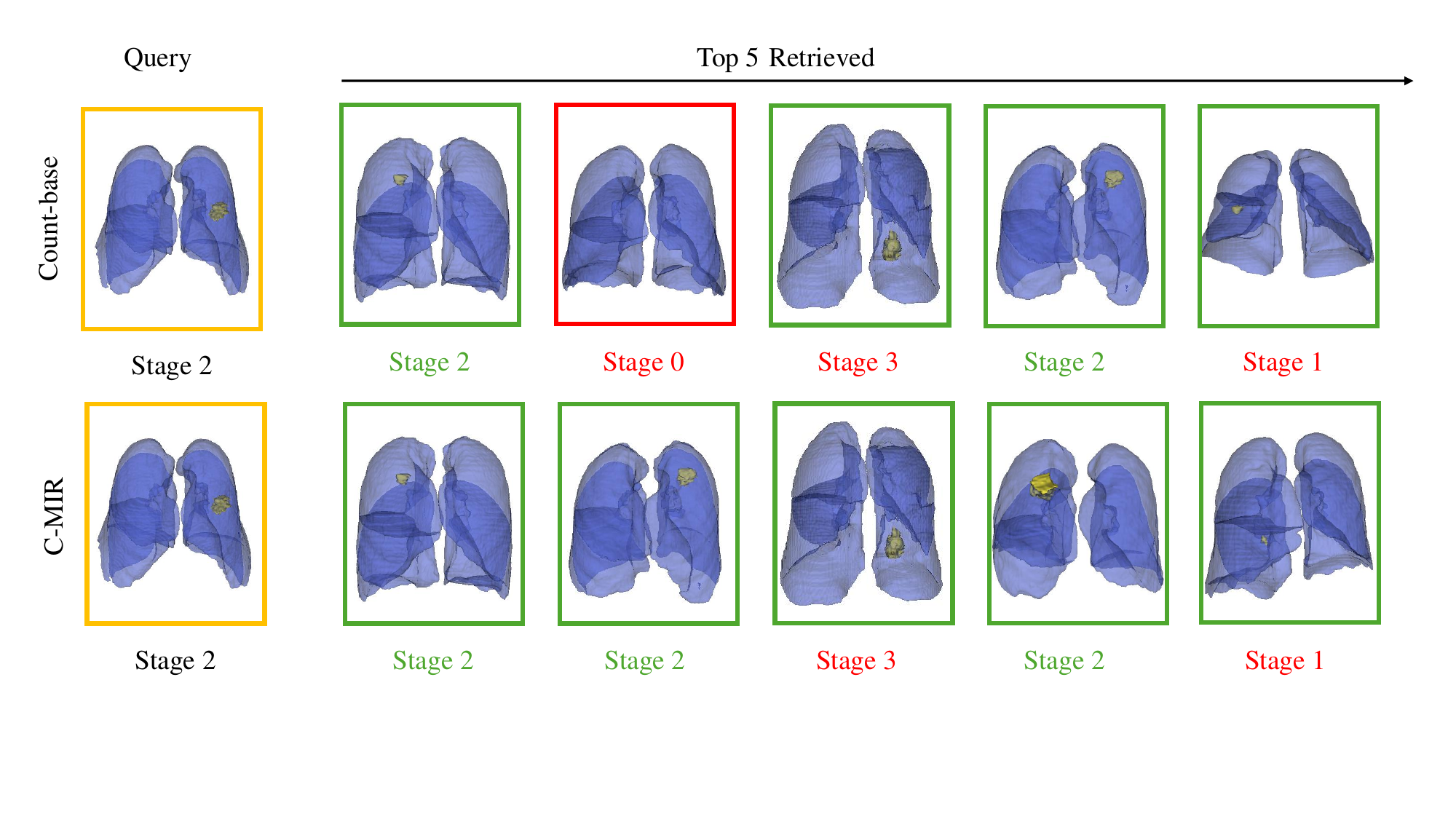}
    \caption{Visual representation of retrieval outcomes for one lung case, with the top five retrieved results. The lung segmentation mask is shown in blue, and the tumors are denoted in yellow. Green boxes indicate accurate tumor flags, while red boxes indicate failures. Stage numbers below each instance show the actual and matched stages, with green for correct and red for incorrect matches. C-MIR improved tumor flagging and staging for the top five retrieved cases. }
    \label{fig:quantitative lung}
\end{figure}

\begin{figure}[!htpb]
    \centering
    \includegraphics[width=0.98\textwidth, trim=0 3cm 0 1cm, clip]{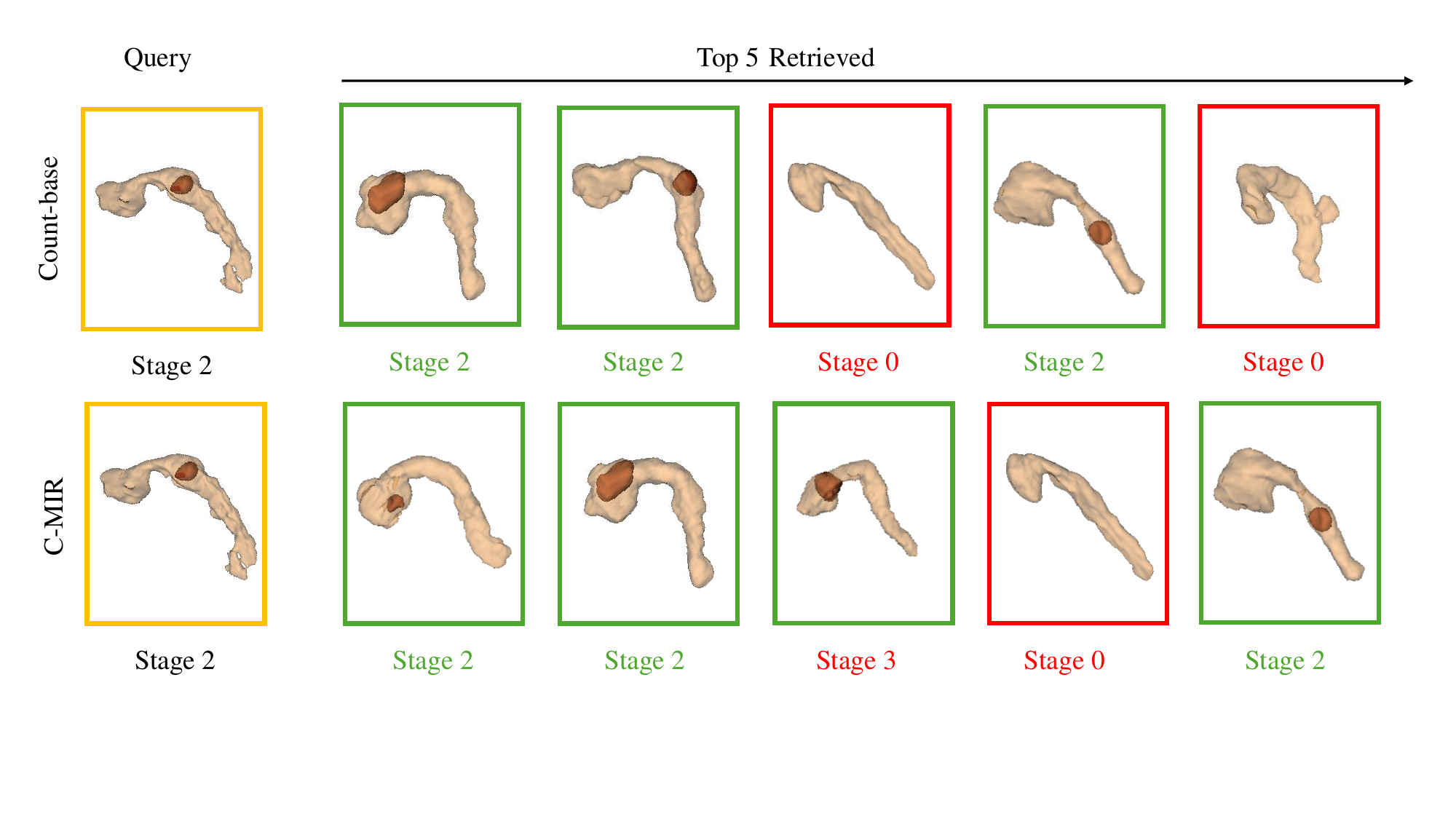}
    \caption{Visual representation of retrieval outcomes for one pancreas case, with the top five retrieved results. The pancreas segmentation mask is shown in yellow and the tumors are denoted in brown. Green boxes indicate accurate tumor flag, while red boxes indicate failures. Stage numbers below each instance show the actual and matched stages, with green for correct and red for incorrect matches. C-MIR improved tumor flagging; however, the tumor staging score remains unchanged for the top five retrieved cases.}
    \label{fig:quantitative pancreas}
\end{figure}

\Cref{fig:quantitative colon}, \Cref{fig:quantitative liver}, \Cref{fig:quantitative lung}, and \Cref{fig:quantitative pancreas} visually depict the top five retrieval outcomes for the colon, liver, lung, and pancreas tumors, utilizing embeddings from the SwinTransformer. The figures compare two retrieval approaches: count-base and C-MIR. The chosen cases illustrate scenarios where C-MIR either enhanced tumor flagging or staging or improved both aspects.
TotalSegmentator model \cite{wasserthal2023totalsegmentator} is used for organ segmentation, while tumor segmentations are obtained from MSD tumor masks \cite{antonelli2022medical}. It is worth mentioning that some organ segmentations are incomplete due to the automatic segmentation process.
In every figure, the query serves as the input image for the search system, and the top five retrieved results are displayed in the same row. The green boxes indicate instances where the tumor flagging was accurate, whereas the red boxes represent instances where tumor flagging was unsuccessful. Below each query or retrieved instance, there is a stage number provided. The stage number shows the actual stage of the query and the corresponding stages of the matched cases. The color of the stages indicates whether the tumor stage is matched correctly, with green for a correct match and red for an incorrect match. The colors of the boxes and stages are independent. For instance, a tumor can be flagged without the correct stage classification, which is indicated by green boxes and red text. It should be noted that the cases presented here were selected from the test set to illustrate common success/failure modes. Full quantitative metrics are reported in \Cref{sec:results and evaluation}.

\Cref{fig:quantitative colon} illustrates a query containing a colon with a stage 4 tumor. The first row shows the top five cases retrieved using the count-base method, with three cases exhibiting a tumor in stages 4, 4, and 3, respectively. Here P@3 and P@5 for tumor flagging is $66\%$ and $60\%$ and P@3 and P@5 for tumor staging is $66\%$ and $40\%$. 
The second row shows the top five cases retrieved using the C-MIR method as re-ranker, with four cases containing a tumor in stages 4, 3, 4 and 4, respectively. Here, P@3 and P@5 for tumor flagging is $100\%$ and $80\%$ and P@3 and P@5 for tumor staging is $66\%$ and $60\%$. This case demonstrates an example where the C-MIR re-ranking improves both tumor flagging and tumor staging.

\Cref{fig:quantitative liver} depicts a query containing a liver with multiple tumors, classified as a stage 3 case according to the count and size of the tumors. The first row displays the top five cases retrieved by the count-base method, all of which have tumors. However, only two of the top retrieved cases contain stage 3 cases and the rest are stage 2. Thus, P@3 and P@5 for tumor flagging is $100\%$ and P@3 and P@5 for tumor staging is $66\%$ and $40\%$. 
The second row shows the top five retrieved cases using the C-MIR method as re-ranker, again with all cases containing tumors. Here, four out of five cases contain tumors of stage 3. As a result, P@3 and P@5 for tumor flagging are $100\%$, and P@3 and P@5 for tumor staging is $66\%$ and $80\%$. This case demonstrates an example where the C-MIR re-ranking improves tumor staging for the top five cases.

\Cref{fig:quantitative lung} demonstrates a query containing a lung with a stage 2 tumor. The first row displays the top five cases retrieved by the count-base method. Four out of five retrieved cases contain tumors of stages 2, 3, 2, and 1. Thus, P@3 and P@5 for tumor flagging is $66\%$ and $80\%$ and P@3 and P@5 for tumor staging is $33\%$ and $40\%$. 
The second row shows the top five retrieved cases using the C-MIR method as re-ranker with all cases containing a tumor. Here, three out of five cases contain tumors of stage 2. Therefore, P@3 and P@5 for tumor flagging is $100\%$ and P@3 and P@5 for tumor staging is $66\%$ and $60\%$. In this case, C-MIR re-ranking improved both tumor flagging and tumor staging for the top five cases.

\Cref{fig:quantitative pancreas} depicts a query containing a pancreas with a stage 2 tumor. 
The first row displays the top five cases retrieved by the count-base method where three out of five retrieved cases contain tumors of stage 2. Thus, P@3 and P@5 for both tumor flagging and staging is $66\%$ and $60\%$. 
The second row shows the top five retrieved cases using the C-MIR method as re-ranker with four out of five cases containing tumors of stages 2, 2, 3, and 2. Therefore, P@3 and P@5 for tumor flagging are $100\%$ and $80\%$ and P@3 and P@5 for tumor staging is $66\%$ and $60\%$. In this case, C-MIR re-ranking improved tumor flagging but the tumor staging score remains the same.

\section{Statistical Analysis: retrieval results for Organ-specific database with/without segmentation}\label{secA2}

\Cref{tab:statistical w/wo}\highlight{ demonstrates the Wilcoxon test on average precision of tumor flagging and tumor staging using the organ-specific database with segmentation versus the organ-specific database without segmentation. The intention of the test is to show whether limiting the search space to organs has a statistically significant impact on the retrieval results for the vanilla aggregation approaches. 
Incorporating segmentation in creating databases significantly influences tumor flagging and staging outcomes, though its impact varies with the choice of image embedding and aggregation method. For tumor flagging, using the organ-specific database with segmentation frequently yielded statistically significant improvements ($p < 0.05$) across multiple organs, particularly when using the BioMedClip embedding (e.g., colon: $p=[0.037-0.049]$ for all aggregation methods; liver: $p=0.020$ for count-base/Sum-Sim). SwinTransformer embeddings also showed significant benefits from using pre-selected slices for flagging of colon and liver tumors ($p=[0.010-0.037]$). In contrast, DreamSim embeddings demonstrated more limited instances of significant improvement in flagging tasks with a similar setup.
For tumor staging, the statistical significance of pre-selecting slices via segmentation of database cases was more specific. Notably, the SwinTransformer embeddings showed a highly significant improvement for colon tumor staging across all aggregation methods ($p=0.010$) and for lung tumor staging ($p=0.014$ for count-based/Sum-Sim). BioMedClip embeddings also showed selective benefits with pre-selected slices (e.g., lung Max-Score: p=0.027; pancreas Max-Score: $p=0.010$). However, using pre-selcted slices for the database via segmentation did not yield statistically significant improvements for staging when using the DreamSim embeddings for any tested organ or aggregation method. These findings underscore that the benefit of the organ-specific database with segmentation is context-dependent, necessitating careful consideration of the embedding model and downstream task.}

\begin{table}[htbp]
  \centering
  \caption{\highlight{Wilcoxon test on average precision of tumor flagging and tumor staging using organ-specific database with segmentation versus organ-specific database without segmentation. The bold-faced values highlight the p-values smaller than 0.05.}  }
  \tiny
    \begin{tabular}{c|c|ccc|ccc}
    \multirow{2}[0]{*}{Organ} & \multirow{2}[0]{*}{Model} & \multicolumn{3}{c}{Flagging p-values w/wo segmentation} & \multicolumn{3}{c}{Staging p-values w/wo segmentation} \\ \cmidrule(l){3-8}
          &       & Count-base & Max-Score & Sum-Sim & Count-base & Max-Sim & Sum-Sim \\ \hline
    \multirow{3}[0]{*}{\begin{sideways}Colon\end{sideways}} & BioMedClip & \textbf{0.049} & \textbf{0.037} & \textbf{0.037} & 0.846 & 0.770 & 0.432 \\
          & DreamSim & 0.375 & \textbf{0.002} & 0.322 & 0.375 & 0.084 & 0.131 \\
          & SwinTrans. & \textbf{0.027} & 0.625 & \textbf{0.037} & \textbf{0.010} & \textbf{0.010} & \textbf{0.010} \\ \hline
    \multirow{3}[0]{*}{\begin{sideways}Liver\end{sideways}} & BioMedClip & \textbf{0.020} & 0.695 & \textbf{0.020} & 0.160 & 0.131 & \textbf{0.037} \\
          & DreamSim & 0.557 & \textbf{0.040} & 0.375 & 0.770 & 0.922 & 0.695 \\
          & SwinTrans. & \textbf{0.027} & 0.846 & \textbf{0.010} & 0.160 & 0.084 & 0.064 \\ \hline
    \multirow{3}[0]{*}{\begin{sideways}Lung\end{sideways}} & BioMedClip & 0.105 & \textbf{0.010} & 0.275 & 0.275 & \textbf{0.027} & 0.432 \\
          & DreamSim & 0.695 & 0.109 & 0.625 & 0.160 & 0.093 & 0.441 \\
          & SwinTrans. & 0.625 & \textbf{0.036} & 0.695 & \textbf{0.014} & 0.432 & \textbf{0.014} \\ \hline
    \multirow{3}[0]{*}{\begin{sideways}Pancreas\end{sideways}} & BioMedClip & \textbf{0.064} & \textbf{0.027} & \textbf{0.010} & 0.131 & \textbf{0.010} & 0.064 \\
          & DreamSim & \textbf{0.064} & 1.000 & \textbf{0.037} & 0.232 & 0.432 & 0.084 \\
          & SwinTrans. & 0.375 & 0.160 & 0.275 & 0.160 & 0.625 & 0.084 \\
    \end{tabular}%
  \label{tab:statistical w/wo}%
\end{table}%

\end{appendices}








\subsection*{Data Availability}
The details of model versions and data splits, including query and database sets, are available upon request. Interested parties can contact the corresponding author for further information on accessing the data.

\section*{Acknowledgement}
The authors would like to thank the Bayer team of the AI Innovation Platform for providing computing infrastructure and technical support.
We thank Timothy Deyer and his RadImageNet team for providing the RadImageNet pre-trained model weights for the SwinTransformer architecture.

\bibliography{sn-bibliography}

\end{document}